\newcommand\SEKIusersusepackages
{
\usepackage{amsmath}
\usepackage{graphicx}
\usepackage{times}
\usepackage{psfrag}
\usepackage{color}
\usepackage{here}
\usepackage[latin1]{inputenc}
\usepackage{xspace}

}

\newcommand\SEKIREVIEWSTATUS{1}

\newcommand\SEKIREVIEWER
{\\Chad E. Brown and Claus-Peter Wirth
\\FR Informatik, Universit\"at des Saarlandes, D--66123 Saarbr\"ucken, Germany
\\E-mail: {\tt cebrown@ags.uni-sb.de}
\\WWW: \url{http://www.ags.uni-sb.de/~cebrown/}
\\
}

\newcommand\SEKIREFEREEPROCESS
{LPAR 2003, 
 10th International Conference on Logic for Programming,
 Artificial Intelligence, and Reasoning;
 full paper, 15 pp. 
}

\newcommand\SEKIDOCUMENTCLASS{0}

\newcommand\SEKIYEAR{2005}

\newcommand\SEKINUMBEROFYEAR{01}

\newcommand\SEKITYPE{1}

\newcommand\SEKIPUBLISHEDTYPE{0}
\newcommand\SEKIPUBLISHEDAS
{Master's thesis, 
 FR Informatik, Universit\"at des Saarlandes, D--66123 Saarbr\"ucken, Germany}
\date
{{\small March 16, 2005\\
\SEKIedition}}

\title{Designing a GUI for Proofs --\\[.3cm]
  Evaluation of an HCI Experiment}
\author
{Martin Homik and Andreas Meier
\\DFKI GmbH, Saarbrücken.
\\\url{{mhomik,ameier}@dfki.de}
}

\def\mippa{MIPPA\xspace}
\def\MULTI{{\sc Multi}\xspace}
\def\activemath{{\sc ActiveMath}\xspace}

\def\OMEGA{$\Omega${\sc mega}\xspace}
\def\loui{{\sc Loui}\xspace}

\def\omega2learn{{\sc TP}4{\sc Learner}\xspace}

\input seki-deckblatt-2
\begin{document}
\makecover
\maketitle
\begin{abstract}Often user interfaces of theorem proving systems focus on assisting particularly trained and skilled users,
  i.e., proof experts.  As a result, the systems are difficult to use for non-expert users. This paper
  describes a paper and pencil HCI experiment, in which (non-expert) students were asked to make suggestions
  for a GUI for an interactive system for mathematical proofs. They had to explain the usage of the GUI by
  applying it to construct a proof sketch for a given theorem. The evaluation of the experiment provides
  insights for the interaction design for non-expert users and the needs and wants of this user group.
\end{abstract}

\section{Introduction}\label{sec:intro}

Human-computer interaction (HCI) is the interdisciplinary study of interaction
between people (users) and computers. Its main goal is making computers more
user-friendly and easier to use. HCI is concerned with methodologies and
processes for designing interfaces, with methods for implementing interfaces, 
with techniques for evaluating and comparing interfaces, with developing new 
interfaces and interaction techniques and with developing descriptive and 
predictive models and theories of interaction \cite{PreeceRogersSharp02}.

More often than not, user interfaces for theorem provers are developed as a
mere add-on to the main proving engine. The result is an interaction design
suitable for proof experts only. As example of such an interaction design
consider the graphical user interface (GUI) {\loui}~\cite{louiJournal99} of
the {\OMEGA} system~\cite{Siekmannetal02} developed in our group in
Saarbr{\"u}cken, Germany: proof presentation as well as the functionalities
provided to the user are tailored to support an expert of the underlying
proving system.

Such an expert-oriented interaction design seems to be sufficient as long as the only users of interactive
theorem provers are trained and skilled users, which are familiar with logic notations and the particularities
of underlying proof engines. However, there are other application domains of theorem provers for which the
user group consists of non-experts. As example domain consider the usage of mathematical systems as cognitive
tools in a learning environment. Currently, the {\mippa} project~\cite{MKM04} develops an interactive learning
system for mathematical proofs, which will be connected with the {\activemath} learning
environment~\cite{activemathAIEDJ01}. The underlying proof engine is the {\MULTI} proof
planner~\cite{MelisMeierCL2000}, which is part of the {\OMEGA} system (see~\cite{MKM04} for the motivation of
using {\MULTI} as proof engine for this task).

In order to make theorem proving systems useful for a larger clientele we have to ask ``What are the needs and
wants of non-expert users?''. In particular, the development of a suitable interaction design is a crucial
point since: {\em Developers often see the functionality of a system as separate from the user interface, with
  the user interface as an add-on. Users, however, do not typically make distinctions between the underlying
  functionality and the way it is presented in the user interface.  To users, the user interface is the
  system.  Therefore, if the user interface is usable, they will see the entire system as usable} (quoted
from~\cite{Dray95}).

In order to develop a system that satisfies the needs and wants of envisioned
users and supports the users in achieving their tasks, interaction design
principles~\cite{PreeceRogersSharp02} suggest an iterative development process
with several interleaved design, experiments, and evaluation phases of the
user interface.  To start with the interaction design for a GUI of an
interactive theorem prover for non-expert users, whose main interest is to
prove a mathematical theorem in an adequate way (whatever the underlying proof
engine or calculus is), we conducted a paper and pencil experiment with
students.  The students had to develop a GUI for an interactive theorem prover
with provided material (paper, pencils of different colors, etc.)  and to use
this GUI to construct a proof sketch for the theorem ``$\sqrt{2}$ is
irrational''.  They were free to invent and use all functionalities of the
assumed underlying proof system they liked. You can find the original exercise
and fotos of the proposals in appendix \ref{app:instructions} and
\ref{sec:propfot}.

This paper describes the experiment and its results. In particular, the evaluation of the student proposals
provides basic insights into wants and needs of non-expert users when interacting with a system for
constructing mathematical proofs.

\vspace*{2ex}
\noindent
In order to avoid some misunderstandings {\em ab initio}:
\begin{itemize}
\item The experiment is not about the evaluation of a given design with users.  Rather, the users were asked
  to freely invent their own proposals.
\item The students represent only one target group of the {\mippa} system: students of computer science,
  mathematics, or engineering. With other target groups (e.g., pupils) similar experiments have to be
  conducted.
\item In an educational context as in the {\mippa} project the main question behind the interaction design
  (and anything else) actually is ``How can learning be supported?'' The context of the experiment and this
  paper, however, is restricted to the more general question of wants and needs of non-expert users.
\end{itemize}

\vfill\pagebreak
\section{The Experiment}\label{sec:setting}

As non-expert users we invited 8 students participating a seminar about proof
planning~\cite{Bundy88,MelisSiekmannAij99} at the computer science department of the Saarland University in
Saarbr\"ucken, Germany. From their studies all students had a profound background in computer science and
mathematics as well as basic knowledge about logic. Because of the seminar the students were also familiar
with notions such as {\em tactic} and {\em method}. Moreover, during the seminar they were briefly introduced
to the {\OMEGA} system and its GUI {\loui}.  In order to stimulate interchange of ideas the 8 students were
grouped into 4 pairs, which are referred to as group A, group B, group C, and group D in the remainder of the
paper.

In the following, we first detail the phases of the experiment and then describe the instructions provided to
the students.

\vfill
\subsection{Sequence of Actions}

The experiment consisted of five phases:

\begin{enumerate}
\item\label{en1} The students were provided with the instruction material (see section~\ref{sec:instr}), which
  they had to read carefully.  Afterwards, they could ask questions. For this phase 15 minutes were scheduled.
\item\label{en2} Each group had to prepare their GUI proposal and a presentation of how to use their
  GUI to prove the example theorem ``$\sqrt{2}$ is irrational''. To do so, the students were supplied with the
  following basic material: paper sheets of different sizes, pencils of different colors, scissors, glue,
  rulers. For this phase 120 minutes were scheduled.
\item\label{en3} Each group had to give a 15 minute presentation of their proposal. All presentations
  were recorded with a video camera.
\item\label{en4} After the presentations the students had another 10 minutes to reflect again their own
  proposal as well as the proposals of the others.
\item\label{en5} In a final discussion the students should point out what elements of the different proposals
  they liked or disliked.
\end{enumerate}

\vfill
\subsection{Instructions}\label{sec:instr}

The following is a brief version of the instructions provided to the students.\footnote{Because of the
  background of the students it was possible to precisely describe their task in the experiment.  For other
  target groups such as pupils without at least some ideas of the construction and arrangement of mathematical
  proofs this might turn out to be a much harder task.}

\vfill\pagebreak

\paragraph{The Example Theorem}

\noindent
The example theorem ``$\sqrt{2}$ is irrational'' was given to the students together with a brief proof sketch:
Assume that $\sqrt{2}$ is rational. Then, there are two integers $n,m$ that satisfy $\sqrt{2}=\frac{n}{m}$ and
that have no common divisor. From $\sqrt{2}=\frac{n}{m}$ follows that $2*m^2=n^2$ (1), which results in the
fact that $n^2$ is even. Then, $n$ is even as well and there is an integer $k$ such that $n=2*k$. The
substitution of $n$ in (1) by $2*k$ results in $2*m^2=4*k^2$ which can be simplified to $m^2=2*k^2$. Hence,
$m^2$ and $m$ are even as well. This is a contradiction to the fact that $n,m$ are supposed to have no common
divisor.

\vfill
\paragraph{The Task}

\noindent
The students were asked to develop a proposal for a GUI for a system to
interactively prove mathematical theorems and to apply this GUI to prove the
given example theorem. In particular, they were instructed to assume an
underlying system that can provide all functionalities they would like to use
to achieve their task. However, they were free to invent and use all
functionalities of the assumed underlying proof system they liked. There were
no instructions or limitations for the design.

For their presentation the students had to use the given material to prepare several states of the GUI, which
cover several states of the proof of the example theorem. To construct the proof they should make use of two
different ways for manipulating a proof under construction: {\em operator-based} proof development and {\em
  island-based} proof development. The former notion means that the system provides operators for proof
manipulation, which can be used during the proof construction. Such an operator has some input and produces
some output, which is introduced into the proof.  For instance, consider an operator for definition unfolding.
When applied to the occurrence of a defined concept such as ``is rational'' in a proof assumption, the
application of the operator derives a new assumption in which the occurrence of ``is rational'' is replaced by
its definition.  Island-based proof development means that the user can freely introduce steps into the proof
by inventing new statements, so called {\em islands} and their relations to other statements in the proof (in
order to indicate from which other statements the island is supposed to follow or which other statements are
supposed to follow from the island).

\vfill
\paragraph{Presentation}

\noindent
We were interested, in particular, in the GUI features and presentations the students would introduce as well
as which functionalities they would demand from the underlying proof system.  Moreover, we were interested in
their motivations and how they would argue for and explain their proposals.  Hence, the students were asked to
point out in their presentations the following points:
\begin{itemize}
\item underlying ideas and the motivation of the group,
\item the organization of their GUI,
\item presentation of proofs and current proof status,  
\item application of operators, 
\item and introduction of island steps.
\end{itemize}

\vfill\pagebreak
\section{Student Proposals and Discussion}\label{sec:proposals}

This section first briefly describes the different proposals presented by the four groups and then gives a
brief account of the discussion following the presentations. The descriptions of the student proposals are
structured wrt.\ the five presentation points introduced in the previous section.

Not surprisingly wrt.\ the free setting of the experiment, many details of the single proposals stay unclear.
Moreover, the experienced reader (i.e., the proof expert) may detect some inconsistencies within the proposals
and the arguments of the students. However, both proposals and arguments provide interesting insights into the
world of thought of the students.

\subsection{Group A: Text-based and Operator-Centered Approach} 

\begin{figure}
  \psfrag{M}[l]{Data} \psfrag{P}[l]{} \psfrag{L}[l]{Lexicon} \psfrag{D}[l]{Database} \psfrag{H}[l]{Help}
  \psfrag{T1}[l]{Theorem} \psfrag{T2}[l]{\colorbox{white}{$\sqrt{2}$ is irrational.}}
  \psfrag{T3}[l]{\underline{Proof:}} \psfrag{T}[l]{Text} \psfrag{F}[l]{Formulae} \psfrag{C}[l]{CaseSplit}
  \psfrag{CS}[l]{Custom Step} \psfrag{CH}[l]{Check} \psfrag{A}[l]{Auto} \psfrag{O1}[c][c]{\small Sugg.}
  \psfrag{O2}[c][c]{\small Op.}  \centering{ \includegraphics[width=.8\textwidth]{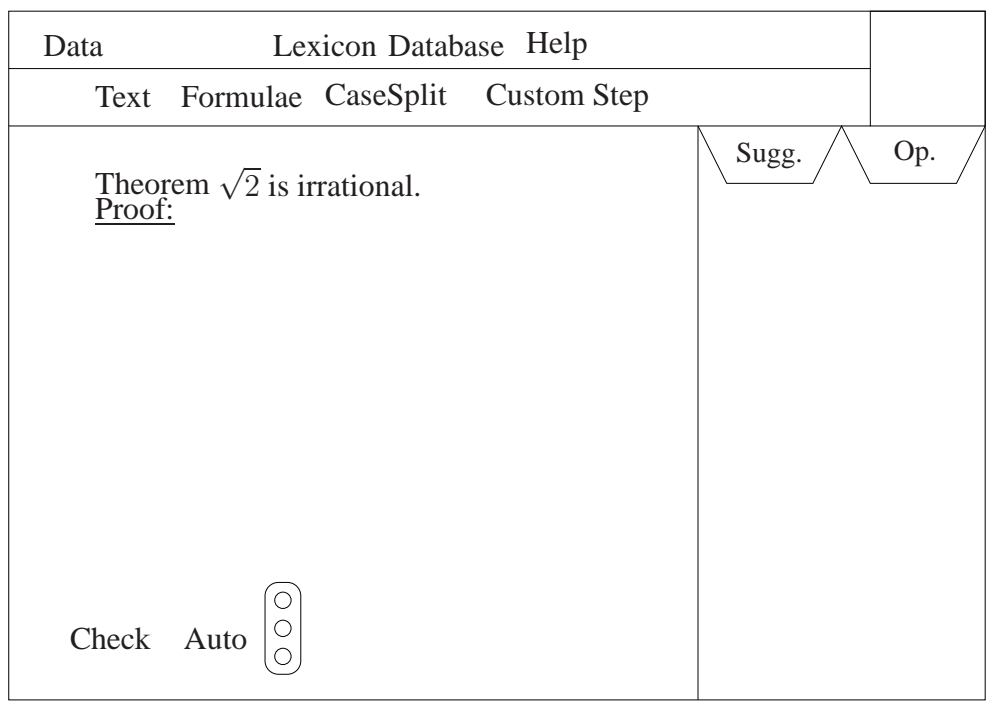} }
  \caption{\label{fig:group1} GUI proposal (Group A).}
\end{figure}

\paragraph{General Idea} The proof is constructed and presented in an accustomed manner ``as taught at
school''.  Therefore, the group argues for a linear proof presentation mixing both text and mathematical
formulae as main presentation format (see Proof Presentation).

\paragraph{GUI Window} The GUI, see Figure~\ref{fig:group1}, is divided into three parts: two bars with menus
and buttons at the top, a presentation area of the current proof state and a choice area for proof techniques
and operators.

The upper bar provides pull-down menus to access standard functionalities such as ``load'' and ``save'' as well as
for lexicon, database, and help access. The lexicon menu allows to browse definitions, theorems or operators.
The database menu allows to retrieve example proofs, which are stored in a database (e.g., proofs of
``similar'' problems). The help menu provides help functions as known from standard software tools.

The lower bar consists of buttons for specific proof manipulations: a custom step button for island
introduction, a case-split button, a text button for the introduction of text parts and a formulae button for
the introduction of formulae.
 
The area for the choice of proof techniques or operators contains a tab for suggestions and a tab for
operators (see Operator Application). 

\paragraph{Proof Presentation} The group favors linear and primarily
text-based proof presentation. At the top are theorem and the assumption
statements. New proof statements are added below independent from whether they
result from forward reasoning from the assumptions or from backward reasoning
from the conclusion. In the statements logical connectives and quantifiers are
omitted and only mathematical formulae ``as known known from school'' are
allowed.  For the example ``$\sqrt{2}$ is irrational'' a presentation looks
like this:
\begin{quote}
\begin{tabbing}
\underline{Theorem:} \= \quad \= The square root of 2 is not rational.\\
\underline{Proof:} \> \> We assume, that the square root of 2 is rational.\\
\> \> Then there exist two numbers $n$ and $m$, being coprime,\\
\> \> \quad such that $\sqrt{2}=\frac{n}{m}$ holds.\\
\> \> \quad Then $2 \times m^2=n^2$ holds.\\
\> \> \ldots
\end{tabbing}
\end{quote}

The underlying system is supposed to continuously check whether the current proof under construction is
correct.  Feedback of this check is provided via a traffic light in the proof presentation area. A green light
shows the checked correctness of the current state, a yellow light the unchecked state. A bright red light
indicates that the underlying system was not able to check the correctness without detecting explicit errors,
whereas a dark red light signs the explicit detection of errors.

Another demanded function of the underlying system is the automated completion of the proof under
construction. This functionality is accessed via an auto button in the proof presentation area. For
automatically closed gaps the completed proof parts should be displayed in the same text-based manner as the
other proof parts.

\begin{figure}
  \psfrag{T1}[l][c]{\textbf{Operator:}}
  \psfrag{T2}[l][c]{\textbf{Indirect}}
  \psfrag{T3}[c][c]{\textbf{?}}
  \psfrag{T4}[c][c]{Apply}
  \psfrag{T5}[c][c]{Cancel}
  \psfrag{T6}[c][c]{Pick}
  \psfrag{T7}[c][c]{Pick}
  \psfrag{T8}[l][c]{Assumptions:}
  \psfrag{T9}[c][c]{Forward}
  \psfrag{T10}[c][c]{Backward}
  \psfrag{T11}[c][c]{\ldots}
  \psfrag{T12}[c][c]{\ldots}
  \centering{ 
    \includegraphics[width=.7\textwidth]{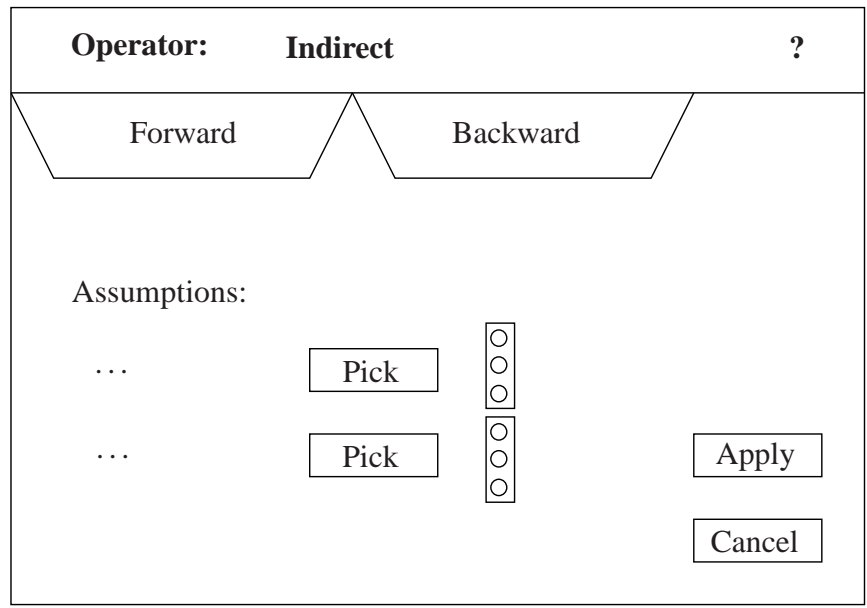}
  }
  \caption{\label{fig:op1} Operator application (Group A).}
\end{figure}

\paragraph{Operator Application} The names of the operators are listed alphabetically in the operator tab.
Clicking an operator name opens a small dialog window showing the pattern of
the input arguments of the selected operator, see Figure~\ref{fig:op1}. The
user marks statements in the proof and drags them into the window to fill the
input arguments. The underlying system immediately checks whether the
arguments are a suitable input for the operator. Feedback of this check is
given in form of traffic lights as well as by error descriptions. The
successful application of an (instantiated) operator results in new text-based
statements in the proof under construction.

To support the application of an operator the underlying system provides so-called {\em suggestions}. When the
user clicks an operator name in the operator tab, then the system computes either full or partial
instantiations of the selected operator. The user can access the computed suggestions via the suggestion tab,
in which the suggestions are ordered wrt.\ the significance the system assigns to them.

\paragraph{Island Introduction} To introduce an island step the user hits the custom step button. This opens a
dialogue window similar to the window for the operator application. The difference is that no operator name
and no pattern is displayed. Rather, the user inputs the island statement via an input editor and determines
the number of input arguments by dragging and dropping as many statements as wanted. Thereby, input statements
are either assumptions or goals, turning the island into a new assumption or goal respectively.

On demand the underlying system checks new islands. That is, the system tries to prove either that an
assumption island follows from the specified assumptions or that the specified goals follow from a goal
island. The system issues feedback on this check to the user.

\paragraph{Miscellaneous} Proof completion might be detected by the underlying proving system automatically or be
indicated explicitely by the user.

\subsection{Group B: Bridge Building Metaphor} 

\begin{figure}
\psfrag{T1}[c][c]{\small Copy}
\psfrag{T5}[c][c]{\small Undo}
\psfrag{T3}[c][c]{\small Redo}

\psfrag{T2}[c][c]{\small Check}
\psfrag{T6}[c][c]{\small Auto}
\psfrag{T4}[c][c]{\small Sugg.}

\psfrag{A}[c][c]{Add Ops}
\psfrag{H}[c][c]{Help}
\psfrag{I}[c][c]{Info}
\psfrag{U1}[c][c]{upper bank}
\psfrag{U2}[c][c]{lower bank}
\centering{ 
 \includegraphics[width=.8\textwidth]{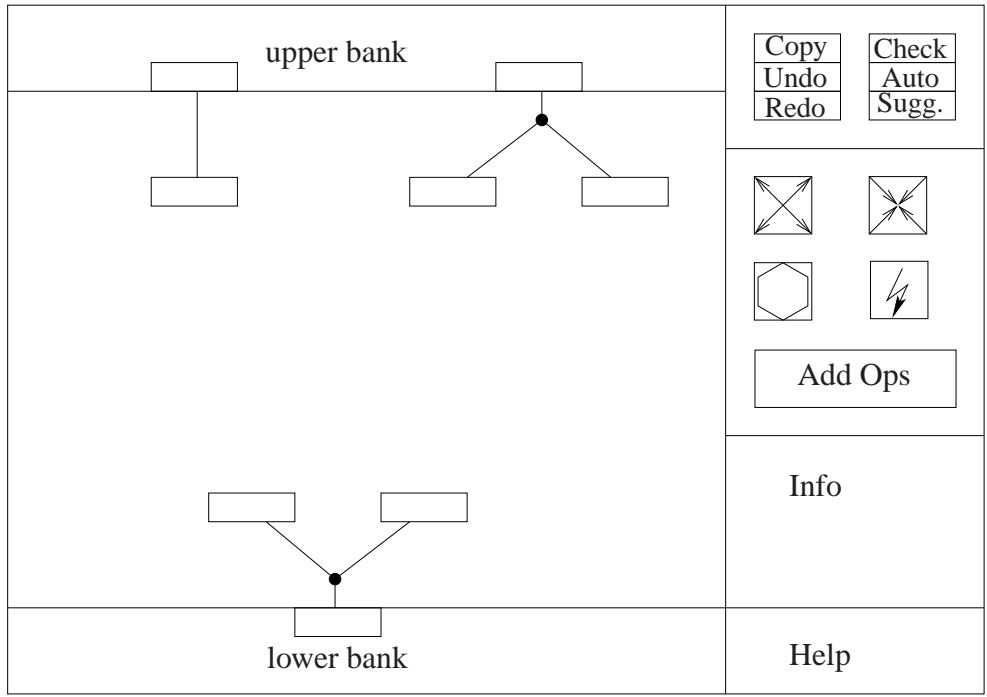}
}
\caption{\label{fig:group2} Bridge Building GUI (Group B).}
\end{figure}

\paragraph{General Idea} The group uses a bridge as metaphor for theorem proving: construct a bridge between
assumptions and conclusion. Moreover, the group invents supporting
visualizations such as icons for operators and drag \& drop techniques for the
bridge construction (see Operator Application).\footnote{The bridge building
  methaphor has already been described by Polya\cite{Polya45} from a pedagogical
  point of view.}

\paragraph{GUI Window} The GUI consists of two parts, see Figure~\ref{fig:group2}. The main area on the left is
used for rendering proofs as a bridge. The area on the right comprises a control and an operator button area
as well as an information and a help area.

The control button area provides ``usual'' control functionalities such as undo, redo, automatic proof
completion and suggestions. Moreover, by hitting a ``Finished'' button, the user can indicate, that he expects
the proof to be finished. In the operator button area most frequently used and relevant (for the current
problem) operators are depicted by icons. The information area shows feedback that the system provides to the
user. The user can request help from the system by dragging objects (e.g., operator icons, buttons, statements
etc.) onto the help field. Then, a feedback window opens and provides specific information on the object.

\begin{figure}[h]
  \psfrag{a}[c][c]{Expansion} \psfrag{b}[c][c]{Collapse} \psfrag{c}[c][c]{Island}
  \psfrag{d}[c][c]{Contradiction} \centering{ \includegraphics[width=.6\textwidth]{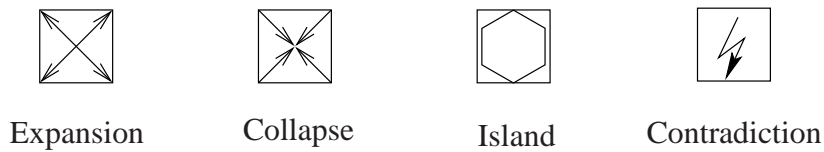} }
  \caption{\label{fig:op2} Operator iconization by Group B.}
\end{figure}

In contrast to other proposals, the group invents icons instead of names for
operator identification.  Each operator is associated with a ``meaningful''
icon, for instance, a lightning bolt may be used for contradiction.  Other
examples are given in Figure~\ref{fig:op2}. ``Expansion'' unfolds the
definition of a selected statement whereas ``Collapse'' has the opposite
effect by contracting an unfolded definition. By hitting the Island button new
islands can be introduced. Experienced users may introduce further icons for
operators.

\paragraph{Proof Presentation} The proof state is presented as bridge under construction between the assumptions
and the conclusion. Bridge nodes consist of nested statements. Edges between the nodes denote operator
applications or island introductions. In the former case, the edge is labeled with the operator icon.

The students pointed out that, if proofs become too large, then some minimization or shrinking mechanism might
be useful, which allows to replace whole proof parts by a single node or label.

\begin{figure}
  \psfrag{T1}[c][c]{\small Copy}
  \psfrag{T5}[c][c]{\small Undo}
  \psfrag{T3}[c][c]{\small Redo}
  
  \psfrag{T2}[c][c]{\small Check}
  \psfrag{T6}[c][c]{\small Auto}
  \psfrag{T4}[c][c]{\small Sugg.}
  
  \psfrag{A}[c][c]{Add Ops}
  \psfrag{H}[c][c]{Help}
  \psfrag{I}[c][c]{Info}
  \psfrag{U1}[c][c]{upper bank}
  \psfrag{U2}[c][c]{lower bank}
  
  \psfrag{DD}[c][c]{\colorbox{white}{Drag\&Drop}}
  \centering{ 
    \includegraphics[width=.8\textwidth]{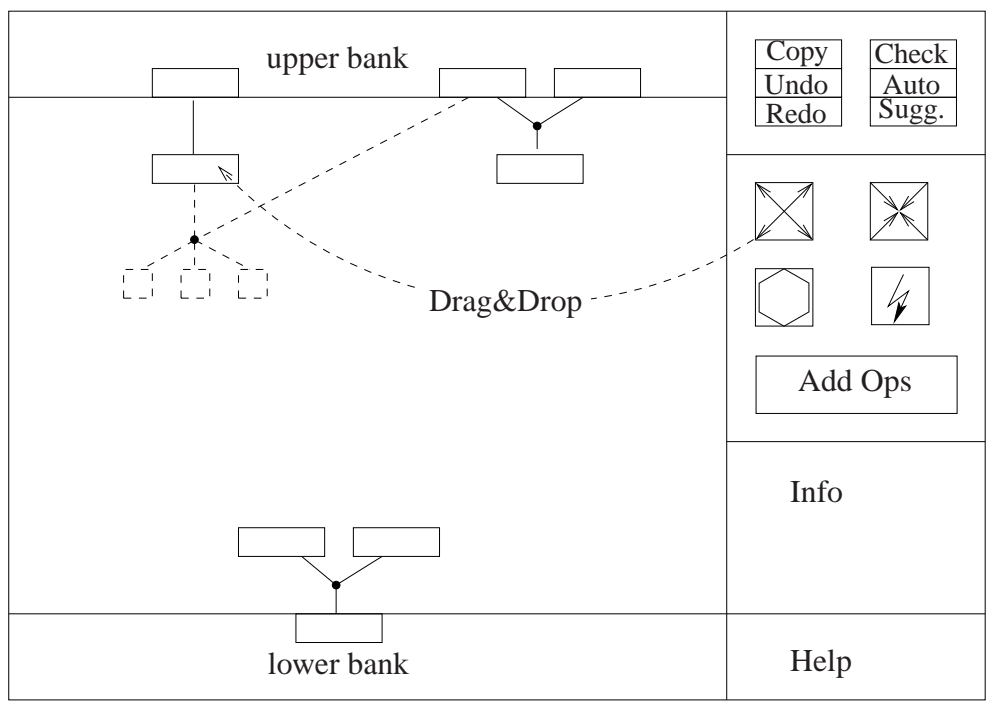}
  }
  \caption{\label{fig:op2b} Operator application (Group B).}
\end{figure}

\paragraph{Operator Application} To apply an operator the user drags the operator icon and drops it onto a proof
statement in the proof state presentation (see Figure~\ref{fig:op2b}). This proof statement is one of the
inputs for the operator. Next, new nodes are inserted into the bridge, which represent the output of the
operator. The new nodes become connected by an edge with the initial statement. If more input is necessary to
apply the operator anchors appear at the new link that the user has to drag \& drop onto further input
statements.

If the initial drag \& drop of the operator icon onto a statement or the drag \& drop of an anchor onto other
statements causes an error (i.e., if the selected statements are not suitable as input for the operator), then
the system provides error feedback in the information field.

\paragraph{Island Introduction} An island is inserted by dragging the island icon and dropping it ``somewhere
into the river''.  Using an input editor the user specifies the island statement, which appears ``in the
river'' as a new bridge node. This new node is initially not connected to any other nodes.  The user connects
it by dragging and dropping anchors that appear at the new island node.

Note that the island approach of group B conceptually differs from the island approach introduced by group A.
Group A allows only  islands that are directly connected with statements already in the proof. Actually,
this results in headlands instead of real islands. As opposed thereto, the approach of group B allows for real
islands, since the island statements do not have to be connected immediately with other statements. 

\paragraph{Miscellaneous} The group suggested more features such as copying of proof parts and automatization of
primitive steps. Moreover, a dialog in the beginning should ask the user about the general proof technique to
employ, for instance, a direct, an indirect or a proof by induction. 

\subsection{Group C: Masking Operator Names}

\begin{figure}
  \psfrag{F}[l]{\colorbox{white}{File}}
  \psfrag{E}[c][c]{\colorbox{white}{Explanations}}
  \psfrag{S}[c][c]{\colorbox{white}{Operators}}
  \psfrag{P1}[c][c]{\colorbox{white}{Direct}}
  \psfrag{P2}[c][c]{\colorbox{white}{Indirect}}
  \psfrag{P3}[c][c]{\colorbox{white}{Induction}}
  \psfrag{P4}[c][c]{\scriptsize Contradiction!}
  \psfrag{T1}[c][c]{$\sqrt{2}$ is irrational}
  \psfrag{T2}[c][c]{$\sqrt{2}$ is rational}
  \psfrag{T3}[c][c]{$\exists m\exists n:\sqrt{2}=\frac{m}{n}$}
  \psfrag{T4}[c][c]{$m,n$ are coprime}
  \centering{
    \includegraphics[width=.8\textwidth]{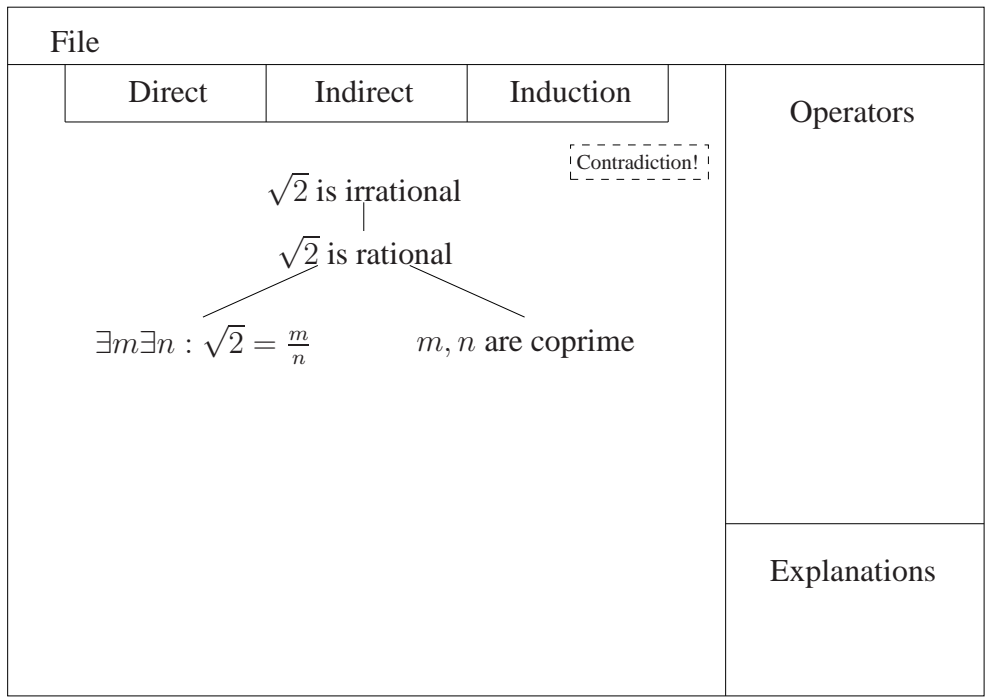}
  }
  \caption{\label{fig:group3} Masking operator names (Group C).}
\end{figure}

\paragraph{General Idea} This group favors a top-down style for proof presentation ``as taught at school or
university''. The proposal differs from other proposals by its omission of operator names.  Operators are
expressed by their input and output (see Operator Application), since ``the main goal is not to learn some
operator names or icons but to understand what follows from what''.

\paragraph{GUI Window} The GUI (see Figure \ref{fig:group3}) has a bar with pull-down menus at the top providing
functionalities known from standard software such as loading and saving of
proofs etc. The rest of the window is divided into two parts. The left part
presents the proof state whereas the right part consists of a field for
operator suggestions and a field for explanations of operator applications
(the Operator Application).

On the top of the proof presentation part is a bar with buttons for ``standard proof techniques'' such as
direct or indirect proof. This bar is built-up dynamically. For instance, when the user decides for an
indirect proof, a contradiction button is added to the bar. The meaning of this new button is that the user
can finish the proof by specifying two contradictory statements in the proof under construction.

\paragraph{Proof Presentation.} Proofs are presented as trees of statements. However, the proof tree is
constructed in a top-down manner similar to the approach of group A. New statements are introduced as nodes
below the statements from which they are derived and connected by edges. However, such edges do not denote
that certain statements are supposed to be logical consequences of other statements. Rather, they are
``storytellers'' of the kind: ``next do $\ldots$ to get $\ldots$''. Assumptions and goals are not explicitly
separated in the tree.

\paragraph{Operator Application} For operator application group C uses also a suggestion mechanism as introduced
already by group A. When the user clicks a statement in the proof, the underlying system computes suggestions
of operator applications with this statement as input. These suggestions are displayed in the operators field
on the right. Neither names nor icon representations of the operators of the suggestions are displayed.
Rather, the output of the suggestion is displayed. If the user clicks a suggestion (via its output), then
detailed information on the underlying step is given in the explanation field, in particular, an explanation
(justification) for the derivation of the output from the input.

Note that the suggestion concepts of group A and group C differ wrt.\ their initialization. The {\em
  operator-oriented} suggestion mechanism of group A is initiated by selecting an operator for which
suggestions are computed. As opposed thereto, the {\em statement-oriented} suggestion mechanism of group C is
initiated by selecting a statement for which suggestions with different operators can be computed.

\paragraph{Island Introduction} Similar to the proposal of group A also this proposal reduces islands to
headlands.  Headlands are introduced by marking a statement and clicking the add button in the suggestion
field, which opens an input editor to enter a new statement. The new statement is inserted into the proof and
is linked with the initially chosen statement. Further links can be added.

\paragraph{Miscellaneous} The group points out that some simplifications like quantifier elimination should be
done automatically by the system. 

The explanation field can be used in different ways. When the user clicks a suggestion (see above), then an
explanation of the step offered by the suggestion is given. If the user clicks a statement in the proof, then
an explanation of the derivation of the statement from its parent nodes is given.

\subsection{Group D: Structuring with Notebooks} 

\begin{figure}
  \psfrag{T1}[c][c]{Tab 1}
  \psfrag{T2}[c][c]{Tab 2}
  \psfrag{T3}[c][c]{Tab 3}
  \psfrag{T4}[c][c]{Tab 4}
  \psfrag{S}[l][c]{Status:}
  \psfrag{P1}[c]{File}
  \psfrag{P2}[c]{Edit}
  \psfrag{P3}[c]{View}
  \psfrag{P4}[c]{Help}
  \centering{ 
    \includegraphics[width=.8\textwidth]{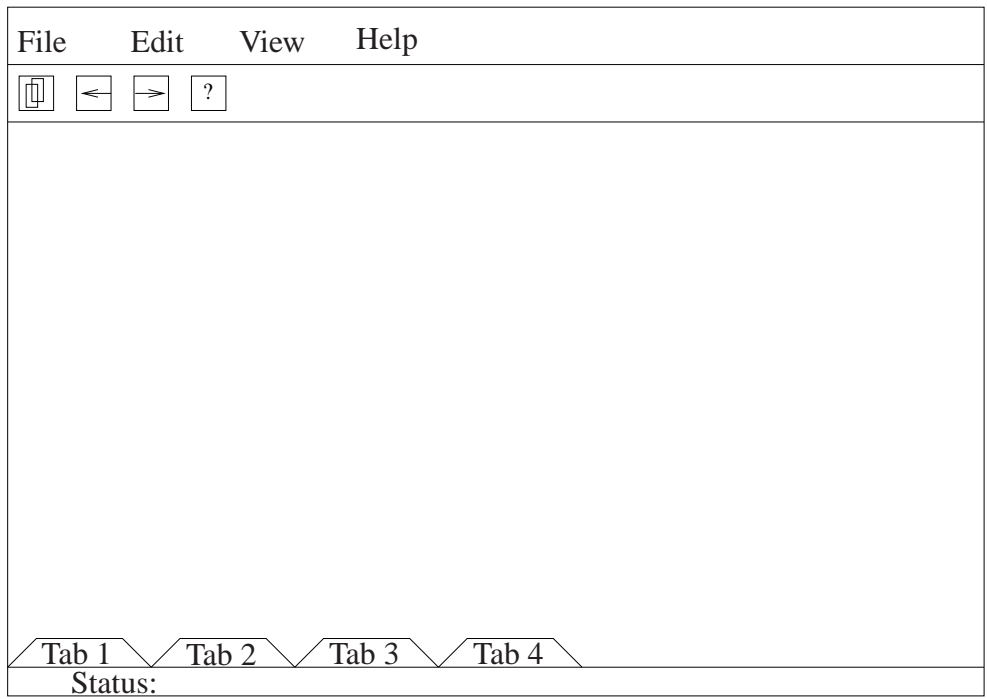}
  }
  \caption{\label{fig:group4} Structuring with Notebooks (Group D).}
\end{figure}

\paragraph{General Idea} The main idea of this approach is to construct and present any proof in the accustomed
pencil and paper manner ``as taught at school''. This implies a linear line of reasoning as well as the
ability of extracting subproofs or conducting several proof attempts. This is realized by a notebook concept,
where each tab corresponds to a subproof. Copying proof parts between tabs should be supported.

\paragraph{GUI Window} The GUI (see Figure~\ref{fig:group4}) has two bars at the top. The upper bar provides
standard pull-down menus ``as known from standard software tools''. The second bar consists of icon buttons
to access standard functionalities such as loading and saving of files, copy and paste of objects, view and
help directly ``as known from standard software tools''. At the bottom of the GUI is a status line for giving
feedback. The main field of the GUI consists of different tabs containing different proof parts or proof
attempts.

\begin{figure}
  \begin{tabular}{|l|} \hline
    \psfrag{T1}[c][c]{\rule{0ex}{3.8ex}$\sqrt{2}$ is irrational} 
    \psfrag{T2}[c][c]{there exist no $m,n\in\cal{Z}$such that $m,n$
      are coprime and $\sqrt{2}=\frac{n}{m}$} \psfrag{T3}[c][c]{We assume, there exist two numbers $n$ and $m$
      in $\cal{Z}$, being coprime, such that $\sqrt{2}=\frac{n}{m}$.}  \psfrag{T4}[c][c]{$2m=n^2$}
    \psfrag{T5}[c][c]{$n^2$ is even} \psfrag{T6}[c][c]{$n$ is even} \psfrag{T7}[c][c]{$\exists k.n=2k$}
    \psfrag{T8}[c][c]{$\exists k.n^2=(2k)^2$} \psfrag{T9}[c][c]{$2m^2=(2k)^2$} \psfrag{T10}[c][c]{\ldots}
    \psfrag{I}[c][c]{$\Updownarrow$}
    \psfrag{U}[c][c]{$\Updownarrow$}
    \psfrag{D}[c][c]{$\Downarrow$}
    \centerline{
      \includegraphics{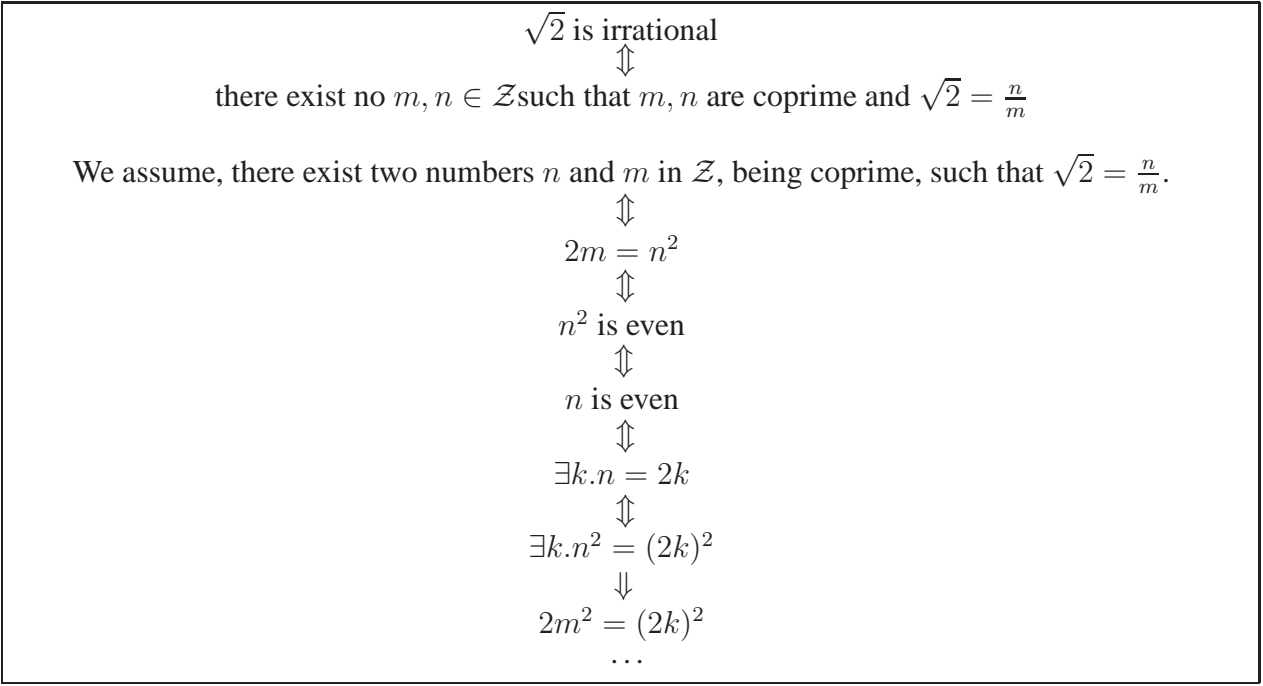}
    }
    \\ \hline
  \end{tabular}
  \caption{\label{fig:team_d_proof} Proof presentation (Group D).}
\end{figure}

\paragraph{Proof presentation.} Proofs are presented in a linear style, where nested statements are connected by
arrows ($\Leftrightarrow, \Leftarrow, \Rightarrow$). The arrows are labeled by operators and denote different
kinds of mathematical consequence relations, e.g., equivalence or implication.  Figure~\ref{fig:team_d_proof}
gives an example presentation of a proof. The students emphasized that statements should be freely
arrangeable and relocatable.

\begin{figure}
  \psfrag{T}[l][c]{\textbf{Operators}}
  \psfrag{T1}[l][c]{\textbf{Recently used:}}
  \psfrag{T2}[l][c]{\textit{Search input}}
  \psfrag{T3}[l][c]{\textbf{Most popular:}}
  \psfrag{T4}[l][c]{- Induction}
  \psfrag{T5}[l][c]{- Indirect}
  \psfrag{T6}[l][c]{- Case split}
  \psfrag{T7}[l][c]{- Substitution}
  \psfrag{T8}[l][c]{- Transitivity}
  
  \centering{
    \includegraphics{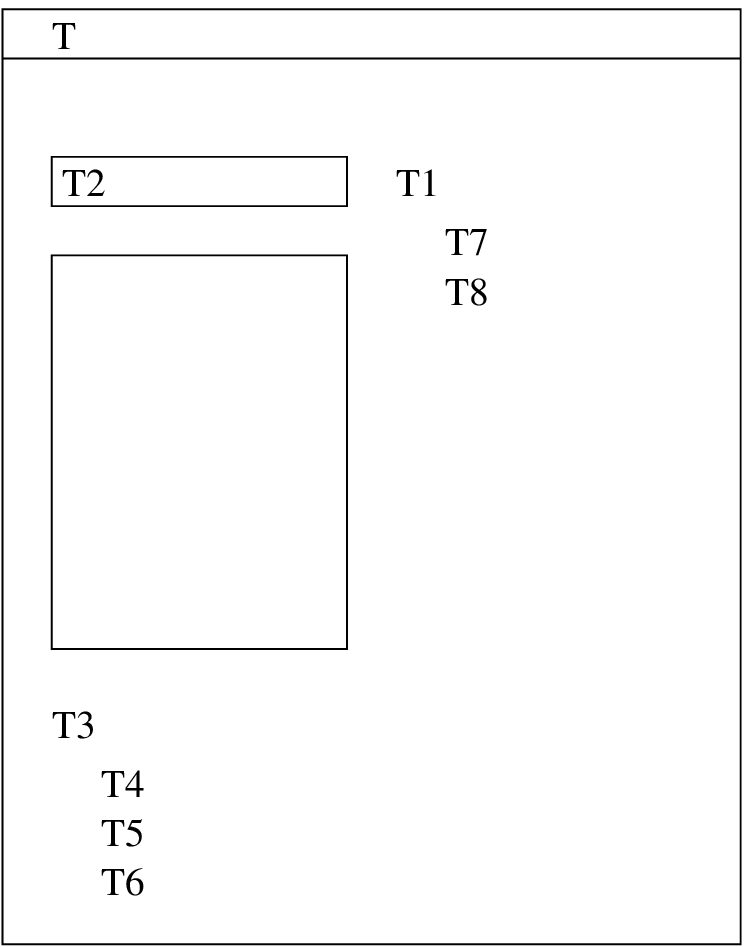}
  }
  \caption{\label{fig:op4} Operator selection (Group D).}
\end{figure}

\paragraph{Operator Application} When the user clicks a statement in the proof, an operator dialog window is
opened (see Figure~\ref{fig:op4}). The window offers lists of most recently used and most popular operators
(e.g., performing induction, indirect proof, or introducing case splits) in which operators are displayed by
their names. When the user selects an operator, then this operator is applied with the initially selected
statement as input. If additional input is necessary, then the user has to provide it by drag \& drop of
further statements into the dialog window.

The user can browse the complete list of available operators. Since this list might be long, supporting search
functionality is provided. This allows for full text search in operator descriptions and for synonyms rather
than pure name-based operator identification. For instance, the search with the inputs \textit{definition
  expansion}, \textit{expansion} or \textit{explanation} results in the operator for definition expansion.

\paragraph{Island Introduction} Similar to group B this approach supports real islands. By clicking somewhere in
the proof development area an island node can be freely located.  The island statement is specified via an
input editor.  When introduced, islands are initially not connected with other statements. The user can freely
introduce connections. The group called such a connection link with an island a ``declaration of intent''.

\subsection{Student Discussion}\label{sec:studdisc}

After reflecting all proposals we asked the students to discuss the benefits and drawbacks of the different
proposals. The following is a summary of the most interesting points. 

\begin{itemize}
\item Members of group A pointed out again the general appropriateness of linear and textual proofs.  However,
  they realized the drawbacks of their proposal wrt.\ island introduction and distinction of assumptions and
  goals.
  
  Similarly, also the members of group D admitted the drawbacks of their linear proof presentation, in
  particular, the lack of separation of assumptions from goals, which both occur at any place in their proofs,
  as well as the distinction of forward and backward reasoning. 
  
  After reflection, all students considered the bridge metaphor invented by group B as more appropriate for
  organizing and overviewing the proof state and progress.
\item All students agreed that the combination of context-sensitive suggestion mechanisms (see group A and C),
  operator filters (most recently, most popular) and full-text synonym search (see group D) would form a
  powerful operator selection widget. Moreover, all students considered freely arrangeable and relocatable
  statements as well as the notebook concept suggested by group D as very helpful.
\item The supervisors mentioned that obviously each group avoided logical notation. The students argued that
  (full) logical notation is not what they are familiar with in mathematical theorem proving from school and
  university.  Moreover, they had objections against dealing explicitly with logical connectives and
  quantifiers in their proofs.
\end{itemize}

\vfill\pagebreak
\section{Evaluation: User Wants and Needs}\label{sec:eval}

As evaluation of the experiment we analyzed the proposals and the discussion of the students for interesting
presentations and features in the GUI as well as for demanded functionalities of the assumed underlying proof
systems. In particular, we were interested in points more or less accepted by all students.

\subsection{Presentations and Features in the GUI}\label{sec:agreedfeat}

The students naturally used many features offered by standard software tools, for instance, pull-down menus,
copy \& paste of content/proof parts and drag \& drop of GUI objects. Such features reflect the experience of
the students with standard software as well as the state-of-the-art in GUI design. More specific, for proof
construction the students suggested and used the following presentations and features.

\paragraph{Nested Statements} Each group used -- at least implicitly -- nested statements composed of text pieces
or formulae.  Sub-statements should be accessible for drag \& drop and operator application.

\paragraph{Text-Based Statements} There was a common agreement about omitting pure logical notation with
quantifiers and connectives. All proposals used some combination of verbalization and mathematical formulae,
such as ``There exists an $x$, such that $\ldots$''.

\paragraph{Input Editor} All groups demanded an input editor for mathematical formulae and composed statements.

\paragraph{Proof State Presentation} Apart from group A, all groups presented proof states as some form of graph
or tree where nodes are labeled with statements and edges are labeled with operators, island connections or
mathematical consequence relations such as $\Leftarrow$, $\Rightarrow$, $\Leftrightarrow$.

In the discussion, the students picked up the bridge construction metaphor very quickly and accepted its
benefits: clear separation between assumptions and goals as well as clear separation between forward reasoning
and backward reasoning.

\paragraph{Notebook} In the discussion, the students agreed that the notebook concept introduced by group D is
useful to structure a complex proof into subproblems or to allow for different proof approaches. They pointed
out, however, that such a concept has to be accompanied by a copy \& paste mechanism for proof parts. 

\paragraph{Operator Identification and Selection} In their presentations all groups introduced special GUI
presentations for operators they considered to be special such as indirect proof, case-split introduction or
induction.

Moreover, the students agreed that an identification of the operator by name only is not appropriate. They
considered iconization of operators as useful but limited approach (assigning every operator a suitable
symbol might be problematic). 

\paragraph{Argument Selection} There was a common agreement that drag \& drop is a necessary support for the
selection of the arguments of an operator. That is, when arguments have to selected for an operator
application, then these arguments should be dragged in the proof presentation and dropped into some operator
application dialog window.

\subsection{Demanded Functionalities of Underlying System}\label{sec:syssupp}

All proposals comprised and demanded many functionalities known from standard software tools such as loading
and saving of documents/proofs, undo function for proof steps and help function (e.g., explanations of the
operators). However, the students also demanded interesting functionalities specific for the construction of
mathematical proofs.

\paragraph{Support for Operator Application} Two of the groups invented some form of context-sensitive
suggestions.  Depending on whether the user marks an operator or a statement for application
(operator-oriented vs.  statement-oriented) the system computes and provides (suggests) completely or
partially instantiated operators. This should free the user from the often laborious task to select all
arguments of an operation application.

\paragraph{Check Operator Arguments} When the user provides the input arguments for an operator, then the system
should check whether the provided input is consistent with the input specification of the operator. If this
check fails, then the system should provide detailed feedback on the reason of the failure.

\paragraph{Check Proof} The system should check the correctness of the proof under construction (either
continously or on demand). It should provide feedback to the user about the result of this check: proof
correct, check failed, errors detected.  In particular, the detection of errors should result in feedback on
both the detected errors and their cause.

\paragraph{Automation Support} 
All groups demanded some automated proof construction support, for instance:
\begin{itemize}
\item The system should perform ``simple'' steps automatically, such that the user does not have to bother
  with them. No group detailed the notion of ``simple''.
\item The system should be able to verify introduced islands. This corresponds to the computation of a
  subproof for the island.
\item The system should be able to complete on demand gaps in the proof under construction automatically. If
  the system fails, it should provide feedback on the reason of the failure.
\end{itemize}

\paragraph{Hints} The system should provide hints on how to proceed, for instance:
\begin{itemize}
\item Depending on the current ``proof strategy'' the system should give general advice. For instance, when
  the user constructs an indirect proof, then the system could provide ``Derive a contradiction!'' as a
  general advice.
\item When providing suggestions, the system should rank these suggestions such that ``the best ones'' are
  ranked first.
\item When a user interaction results in a failure, then the system should offer specific hints and guidance
  how to overcome the failure. For instance, when the user specifies input arguments not suitable for an
  operator, then the system could suggest suitable input arguments.
\end{itemize}

\paragraph{Feedback} All students pointed out that the provision of feedback as response to user interactions is
very important. Feedback should follow both successful user interactions -- confirming the user interaction --
and failing user interactions -- explaining the problem as detailed as possible.

\paragraph{Retrieve Examples from Database} The system should support the retrieval of data from a proof
database.  In particular, it should support the retrieval of proofs for ``similar'' problems as well as the
retrieval of example applications of selected operators. This allows the study of successful proofs and
successful operator applications. None of the groups elaborated the notion of ``similar'' problems.

\vfill\pagebreak
\section{Conclusion}

In this paper, we described a paper and pencil HCI experiment conducted with graduate computer science
students.  The aim of the experiment was to gain basic insights into the wants and needs of non-expert users
of a system for mathematical proof, i.e., users that are not particularly trained wrt.\ a certain theorem
proving system and its functionalities.

In the experiment the students were asked to invent a GUI for an interactive
system for mathematical proof.  They had to explain their GUI structure and
the used functionalities of the assumed underlying system with the example
problem ``$\sqrt{2}$ is irrational''. In a subsequent discussion the students
were asked to reflect their own proposal as well as the other proposals and to
point out features and functionalities they particularly liked or disliked.

\noindent
Among the (more or less) commonly agreed GUI features and useful functionalities are:
\begin{itemize}
\item statement presentation that avoids logical notation and mixes textual presentation with mathematical
  formulae,
\item a graphical presentation of proof states that supports distinguishing goals and assumptions as well as
  forward and backward reasoning,
\item rich system support for performing proof steps (e.g., computation of suggestions), 
\item rich system feedback on user interactions (e.g., explanation of failures), 
\item automated proof construction support (e.g., simple steps should be done automatically),
\item provision of hints (e.g., hints to overcome a failure). 
\end{itemize}
Moreover, the students naturally demanded functionalities known from standard software, e.g., loading and
saving of proofs and availability of an undo function. They also naturally used state-of-the-art GUI
techniques such as copy \& paste and drag \& drop.

An interesting observation was that, initially, a majority of the students
argued for a linear character of proofs, not only of the final proof but also
of the intermediate proof states during the construction. They repeatedly
stated that such a linear form of proof and proof state presentation is their
experience from school and university lectures and tailored their GUI
presentations to this form. This observation is consistent with empirical
studies that suggest that student's deficiencies in their mathematical
competence with respect to understanding and generating proofs are connected
with a basic misunderstanding of the theorem proving process. Typical
presentations of proofs in math books and lectures present proofs as linear
constructs that derive the conclusion from the assumptions. This results in
the shortcoming of learners understanding of theorem proving as a highly
non-linear and hierarchical search and construction process.  Therefore it is
not surprising, that the students picked up the bridge building metaphor as a
more adequate way of proof construction, since it enforces and clarifies
proof states and proof reasoning. 
 
Based on the observed user wants and needs we will develop a GUI prototype. In
order of achieving usability, especially for our non-experts target group, we
intend to evaluate our prototype theoretically according to the cognitive
dimensions framework \cite{conf/ct/BlackwellBCGGKKLNPRRWY01} as well as
experimentally with target users. To focus on our concrete task in the
{\mippa} project (see section~\ref{sec:intro}) we will also conduct further
experiments explicitly focusing on an interactive system for learning
mathematical proof.

\paragraph{Acknowledgments} 
We would like to thank Paul Cairns, Antonio Kr{\"u}er and Erica Melis for stimulating discussions on the
experiment and Chad E.  Brown for feedback on the paper.


\newpage
\begin{appendix}

\section{Original German Instructions for the Students}\label{app:instructions}

\begin{center}
  {\LARGE Seminar Experiment\\[1ex]
    16.07.2004}
\end{center} 

\begin{center}
  Hallo Liebe Studenten und Willkommen zu unserem Experiment!\\[2ex]
\end{center}

\vspace*{4ex}
{\Large {\bf 1. Einleitung}}

\vspace*{2ex}
{\bf Um was geht es hier und heute?}

Ihr habt flei{\ss}ig an unserem Proof Planning Seminar teilgenommen und auch
ein wenig mit {\OMEGA} und seinem GUI LOUI gearbeitet. Nun wollen wir etwas
sehr Schwieriges von euch: Ihr sollt f{\"u}r die n{\"a}chsten Stunden alles
vergessen, was ihr {\"u}ber Proof Planning, {\OMEGA} und LOUI wisst. Denn wir
wollen mit euch ein Experiment durchf{\"u}hren an Hand dessen Resultate wir
ein Graphical User Interface entwickeln wollen mit dem Studenten mathematische
Beweise f{\"u}hren sollen. Ein dahinter liegendes Beweis-Tool soll Support und
Funktionalit{\"a}ten zur Verf{\"u}gung stellen. Aber das GUI soll auf keinen
Fall speziell f{\"u}r {\OMEGA} entwickelt werden. Daher sollt ihr eben nicht
von euren {\OMEGA} Erfahrungen ausgehen, sondern sollt eurer Kreativit{\"a}t
freien Lauf lassen und ein GUI sowie eine Beweisdarstellung nach eurem
Geschmack entwerfen.  (M{\"u}ssen mathematische Aussagen in Formeln kodiert
sein? Muss ein Beweis als Baum dargestellt werden? Was hat euch an LOUI und
{\OMEGA} besonders gest{\"o}rt? Wie w{\"u}rdet ihr gerne Beweise aufbauen und
manipulieren?).

\vspace*{2ex}
{\bf Wie soll das Ganze ungef{\"a}hr ablaufen?}

Wir geben euch eine Beweisaufgabe vor, die in einer bestimmten Art und Weise
bearbeitet werden soll.  Desweiteren geben wir euch Material vor: Stifte,
Papier, Kleber, Lineale sowie vorgefertigte GUI Elemente wie verschiedene
Arten von Windows und Menues. Ihr werdet in Zweier-Gruppen eingeteilt und
sollt dann jeweils ein GUI entwerfen um damit die beschriebene Aufgabe zu
bearbeiten. GUI entwerfen hei{\ss}t: Nehmt vorgefertigte GUI Elemente oder
entwickelt und fertigt euch selbst GUI Elemente, arrangiert diese auf den
ausgegebenen DIN A2 Bl{\"a}ttern und spielt dann durch, wie die Aufgabe in
eurem GUI abl{\"a}uft. F{\"u}r diese Konzeptions- und Vorbereitungsphase habt
Ihr ca. 70 Minuten Zeit. Anschlie{\ss}end stellt jede Gruppe ihren Ansatz am
Flipchart vor (jeweils bis zu 15 Minuten) und wir diskutieren gemeinsam die
Ans{\"a}tze.

\vspace*{4ex}
{\Large {\bf 2. Material}}

\noindent
Als Basismaterial stehen euch zur Verf{\"u}gung:

\begin{itemize}
\item DIN A2 Bl{\"a}tter zum Arrangieren des GUI's
\item Stifte in verschiedenen Farben
\item Kleber
\item Schere
\item Lineal
\item DIN A3 und DIN A4 Bl{\"a}tter (um euch beliebige GUI Elemente selbst daraus zu schneidern)
\end{itemize}

\noindent
Folgende vorgefertigten GUI Elemente liefern wir euch schon mal mit:

\begin{itemize}
\item verschiedene Windows:
  \begin{itemize}
  \item Standard Leer-Window (erg{\"a}nzt, was immer ihr wollt)
  \item Listen Windows
  \item Text Windows
  \item Tabellen Windows
  \item Dialog Windows
  \end{itemize}
\item Kontext Menues
\end{itemize}

{\bf Hinweis: Das gesamte vorgegebene Material ist eigentlich nur dazu da euch
  Impulse zu liefern. Wenn euch das vorgegebene Material nicht gef{\"a}llt
  bzw. es nicht ausreicht um eure Vorstellungen zu realisieren, dann d{\"u}rft
  ihr euch selbst aus dem Basismaterial basteln, was immer ihr wollt!}

\vspace*{4ex}
{\Large {\bf 3. Aufgabe}}

\noindent
{\bf Die Beweisaufgabe, die Ihr bearbeiten sollt ist:}

\begin{description}
\item[Theorem:] Die Wurzel von 2 ist nicht rational.
\item[Beweisskizze:] Nehmen wir an, dass $\sqrt{2}$ rational ist. Dann gibt es
  zwei ganze Zahlen $n,m$, die teilerfremd sind, so dass
  $\sqrt{2}=\frac{n}{m}$. Dann gilt auch $2*m^2=n^2$ (1), womit $n^2$ eine
  gerade Zahl ist. Nach dem Satz {\"u}ber die eindeutige
  Primzahlfaktorisierung ganzer Zahlen muss dann auch $n$ gerade sein.  Damit
  gilt: $n=2*k$ (2) f{\"u}r eine ganze Zahl $k$. (2) eingesetzt in (1) ergibt
  $2*m^2=4*k^2$ was sich zu $m^2=2*k^2$ k{\"u}rzen l{\"a}sst. Damit sind dann
  auch $m^2$ und $m$ gerade
  Zahlen. Somit haben wir, dass sowohl $n$ als auch $m$ gerade sind. Dies ist ein Wiederspruch dazu, dass $n,m$ teilerfremd sind.\\[3ex]
\end{description} 

\noindent
{\bf Was genau ist zu tun?}

Ihr sollt die schrittweise Erstellung dieses Beweises durchspielen und dabei einige Zust{\"a}nde eures GUI
vorbereiten. Wir unterscheiden dabei explizit zwischen GUI Zustand und Interaktion. Der schrittweise
Beweisaufbau besteht aus einer wechselnden Folge von GUI Zustand und Interaktion. Jeder GUI Zustand soll auf
einem extra DIN A2 Papier vorbereitet werden. Die anschlie{\ss}ende Interaktion soll dann an diesem DIN A2
Papier durchgef{\"u}hrt werden. Dazu muss dann auch entsprechendes Material vorbereitet werden, das ihr bei
der Vorf{\"u}hrung auf das DIN A2 Papier pinnen k{\"o}nnt (alles wird
an Korkwand vorgef{\"u}hrt) um eure Interaktionen darlegen zu k{\"o}nnen.\\[3ex]

\noindent
{\bf Allgemeiner Hinweis: Beweismanipulation}

Es gibt allgemein zwei M{\"o}glichkeiten einen Beweis zu manipulieren:
\begin{enumerate}
\item Das System stellt Operatoren zur Verf{\"u}gung, die angewandt werden
  k{\"o}nnen (wir vermeiden jetzt mal den Begriff Methode oder Taktik). Dieser
  Operator bekommt einen bestimmten Input und liefert dann dazu einen
  bestimmten Output. Zum Beispiel der Operator {\sc DefinitionExpansion} wird
  auf das Vorkommen eines definierten Konzeptes in einer Aussage angewand um
  das definierte Konzept zu expandieren. D.h. er bekommt als Input das
  konkrete Vorkommen des Konzeptes in einem Ausdruck sowie die Definition des
  Konzeptes. Er liefert als Output den entsprechenden Ausdruck, in dem dieses
  Vorkommen des Konzeptes expandiert ist.
\item Unabh{\"a}ngig von Operatoren k{\"o}nnen auch jederzeit Beweisteile skizziert werden, indem Aussagen
  frei eingef{\"u}gt und verlinkt werden. Etwa wie folgt:
  \begin{itemize}
  \item (vorw{\"a}rts) Ich f{\"u}ge Aussage $A$ in meinem Beweis ein und $A$ soll folgen aus bereits gegebenen
    Aussagen $A_1$ und $A_2$.
  \item (r{\"u}ckw{\"a}rts) Ich muss Aussage $A$ zeigen und $A$ soll folgen aus neuen Aussagen $A_1$ und
    $A_2$, die ich in meinen Beweis einf{\"u}ge.
  \end{itemize}
\end{enumerate}

Ihr solltet davon ausgehen, dass immer eine Vielzahl von Operatoren zur
Verf{\"u}gung stehen (entsprechend einer gr{\"o}{\ss}eren Menge von Methoden
oder Taktiken). Es kommt aber vor, dass eventuell keine Operatoren da sind,
die jetzt genau die Schritte aus-f{\"u}hren, die ihr gerne h{\"a}ttet. Dann
k{\"o}nnt ihr eure
Schritte (d.h. euren Beweis) nur {\"u}ber frei eingegebene Aussagen ausdr{\"u}cken.\\[3ex]

\noindent
{\bf Allgemeiner Hinweis: Funktionalit{\"a}ten des Beweissystems}

Das darunterliegende Beweissystem stellt eine Menge von
Basis-Funktionalit{\"a}ten zur Verf{\"u}gung, z.B.:
\begin{itemize}
\item Operator anwenden (sofern anwendbar)
\item Test von Anwendbarkeit von Operator
\item Generierung von Vorschl{\"a}gen (= Liste von anwendbare Operatoren)
\item Instantiierung von Variablen
\item Einf{\"u}gen und Verlinken von Aussagen (ohne Operatoren) (beinhaltet z.B. auch den Check ob Zykel
  entstehen)
\item Angabe des Beweiszustandes (Was f{\"u}r Ziele? Was f{\"u}r Annahmen? Was f{\"u}r Annahmen relevant
  f{\"u}r welche Ziele? $\ldots$)
\item $\ldots$
\end{itemize}

{\bf Hinweis: Diese Liste der Funktionalit{\"a}ten ist nicht exklusiv. Es
  steht euch frei euch noch beliebige weitere Funktionalit{\"a}ten zu
  {\"u}berlegen und einzusetzen (z.B. Anfrage an Auto-Mode: Was w{\"u}rdest Du
  als n{\"a}chstes machen? etc.). Ihr d{\"u}rft auch hier kreativ sein!}

\vspace*{4ex}
{\Large {\bf 4. Schrittweise Erstellung des Beweises}}

Folgenden Beweisaufbau sollt ihr nun konkret durchdenken und entsprechendes
Material f{\"u}r die Pr{\"a}sentation vorbereiten.

\begin{description}
\item[GUI Zustand 1:] Das Problem ``$\sqrt{2}$ ist nicht rational'' wurde
  geladen. Wie sieht euer GUI aus?
\item[Interaktion 1:] Ihr entschlie{\ss}t euch einen indirekten Beweis zu
  f{\"u}hren. Daf{\"u}r steht der Operator {\sc Indirekt} zur Verf{\"u}gung.
  Wie wollt ihr diesen Operator im GUI anwenden? Die Anwendung dieses
  Operators gelingt. Was soll der GUI f{\"u}r Feedback geben?
\item[GUI Zustand 2:] F{\"u}r den indirekten Beweis muss angenommen werden, dass $\sqrt{2}$ rational ist um dann 
  anschlie{\ss}end einen Wiederspruch herzuleiten. Wie sieht euer GUI aus?
\item[Interaktion 2:] {\em x ist-rational} ist ein definiertes Konzept. Die
  Definition sagt, dass es f{\"u}r x eine Darstellung als Bruch zweier
  teilerfremden ganzen Zahlen gibt. Ihr wollt diese Definition aus einer
  angeschlossenen Datenbank erfragen. Wie wollt ihr das im GUI machen?  Was
  soll der GUI an Output oder weiterem Feedback liefern?
  
  Ihr versucht (fehlerhafterweise) diese Definition mittels des Operators {\sc DefinitionExpansion} auf
  $\sqrt{2}$ anzuwenden. Wie wollt ihr das im GUI machen?  Diese Anwendung des Operators schl{\"a}gt fehl. Was
  soll der GUI an Output oder weiterem Feedback liefern?
  
  Anschlie{\ss}end wendet ihr diese Definition auf {\em $\sqrt{2}$
    ist-rational} an (das in irgendeiner Form in eurem Beweis vorkommen muss).
  Wie wollt ihr das im GUI machen?  Diese Anwendung des Operators gelingt.
  Was soll der GUI f{\"u}r Feedback geben.
\item[GUI Zustand 3:] Das Resultat dieser Definition Expansion muss irgendwie
  $\sqrt{2}=\frac{n}{m}$ enthalten. Wie sieht euer GUI aus?
\item[Interaktion 3:] Ihr wollt $2*m^2=n^2$ als frei spezifizierte Aussage
  einf{\"u}hren und spezifizieren, dass es aus $\sqrt{2}=\frac{n}{m}$ folgen
  soll. Wie wollt ihr das im GUI machen?  Was soll der GUI an Feedback
  liefern?
\item[GUI Zustand 4:] Anschlie{\ss}end leitet ihr in einer Folge von frei
  spezifizierten (aber verlinkten) Aussagen ab, dass sowohl $n$ alsauch $m$
  gerade sind. Wie sieht euer GUI aus?
\item[Interaktion 4:] Ihr leitet den Wiederspruch her aus den (inzwischen in
  irgendeiner Form enthaltenen) Aussagen, dass $n$ und $m$ gerade sind sowie,
  dass $n$ und $m$ teilerfremd sein sollen. Dies schlie{\ss}t
  euren Beweis. Wie wollt ihr das im GUI machen? Was soll der GUI an Feedback liefern?\\[3ex]
\end{description}

{\bf Hinweis: Erst ein GUI Gesamtkonzept {\"u}berlegen, in dem sich diese
  Sachen realisieren lassen. Dann erst anfangen zu basteln!}

\vspace*{4ex}
{\Large {\bf 5. Pr{\"a}sentation}}
  
Bei der Pr{\"a}sentation soll neben dem Durchspielen des Beweises insbesondere
Folgendes erkl{\"a}rt werden:

\begin{itemize}
\item Was sind die grundlegenden Elemente des GUI's?
\item Was bedeuten diese Elemente?
\item Wie und wo wird der momentane Beweiszustand dargestellt? 
\item Wie und wann werden Funktionalit{\"a}ten des darunterliegenden
  Beweissystems benutzt und wie werden diese Funktionalit{\"a}ten {\"u}ber die
  GUI Elemente angesprochen?
\end{itemize}

\vspace*{3cm}
\begin{center}
  {\bf Noch Fragen?}
\end{center}

{\bf Um es nochmal zu sagen: Nicht an {\OMEGA} Erfahrung anlehnen (oder an
  irgend ein anderes Tool, das ihr kennt). Lasst eurer Kreativit{\"a}t freien
  Lauf!}

\begin{center}
  {\bf Viel Spa{\ss}!}
\end{center}

 \newpage
 \section{Proposal Fotos}\label{sec:propfot}
 We have attached some fotos, taken during the presentation of each aproach.
 They essentially show the structure of a GUI window, a proof presentation and
 the application of operators.

  \begin{figure}[h] 
  \centering{ 
   \includegraphics[width=.8\textwidth]{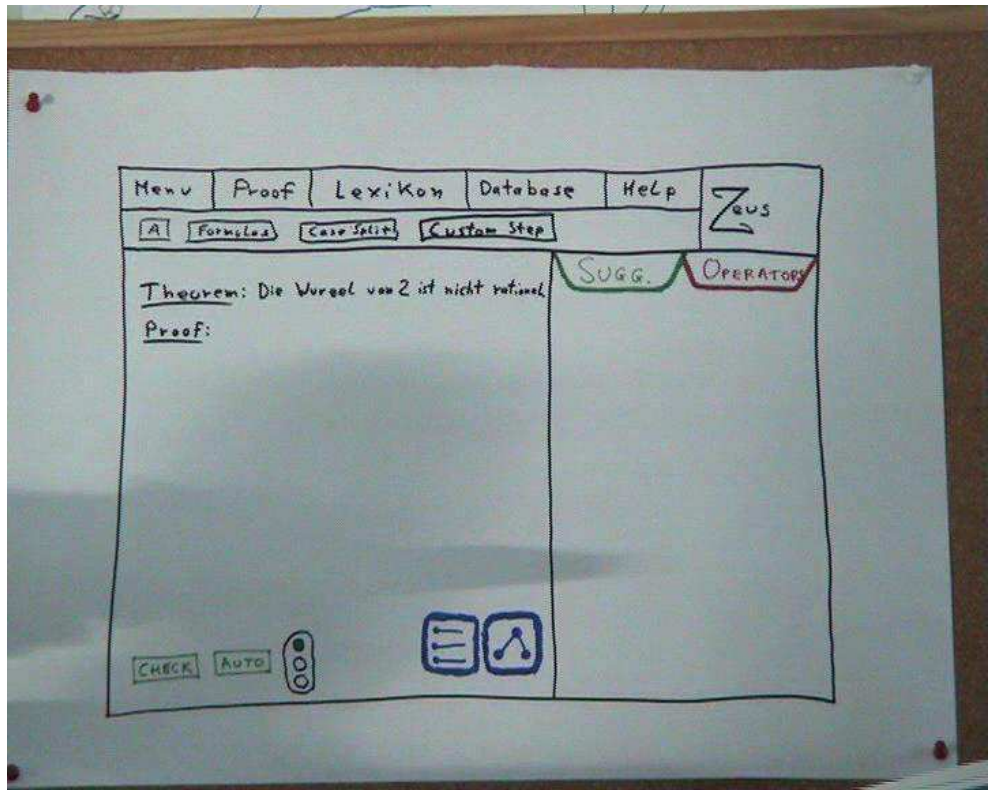}
  }
  \caption{\label{fig:team_a_window} GUI window (Group A)}
  \end{figure}

  \begin{figure} 
  \centering{ 
   \includegraphics[width=.8\textwidth]{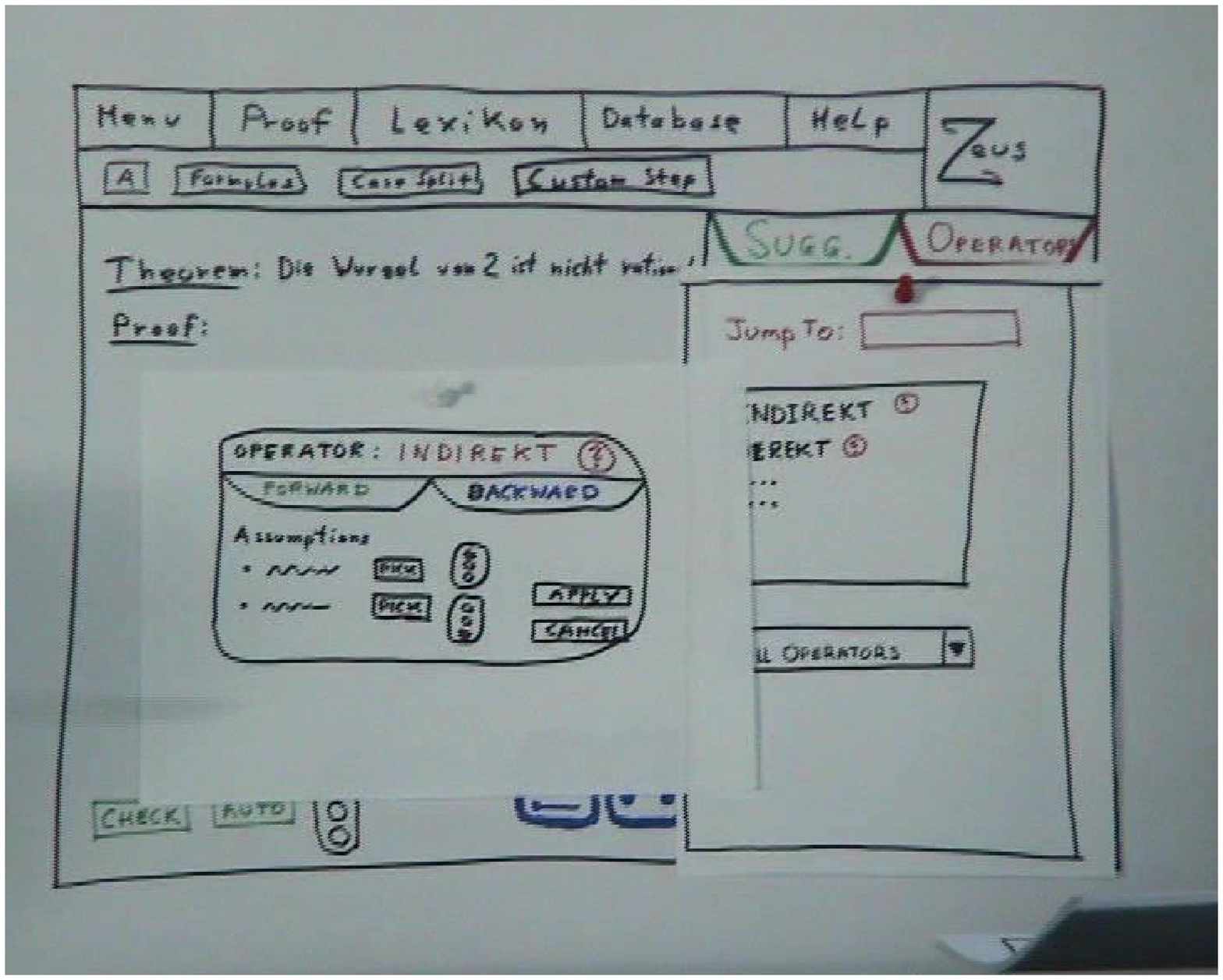}
  }
  \caption{\label{fig:team_a_operator} Operator application (Group A)}
  \end{figure}

  \begin{figure} 
  \centering{ 
   \includegraphics[width=.8\textwidth]{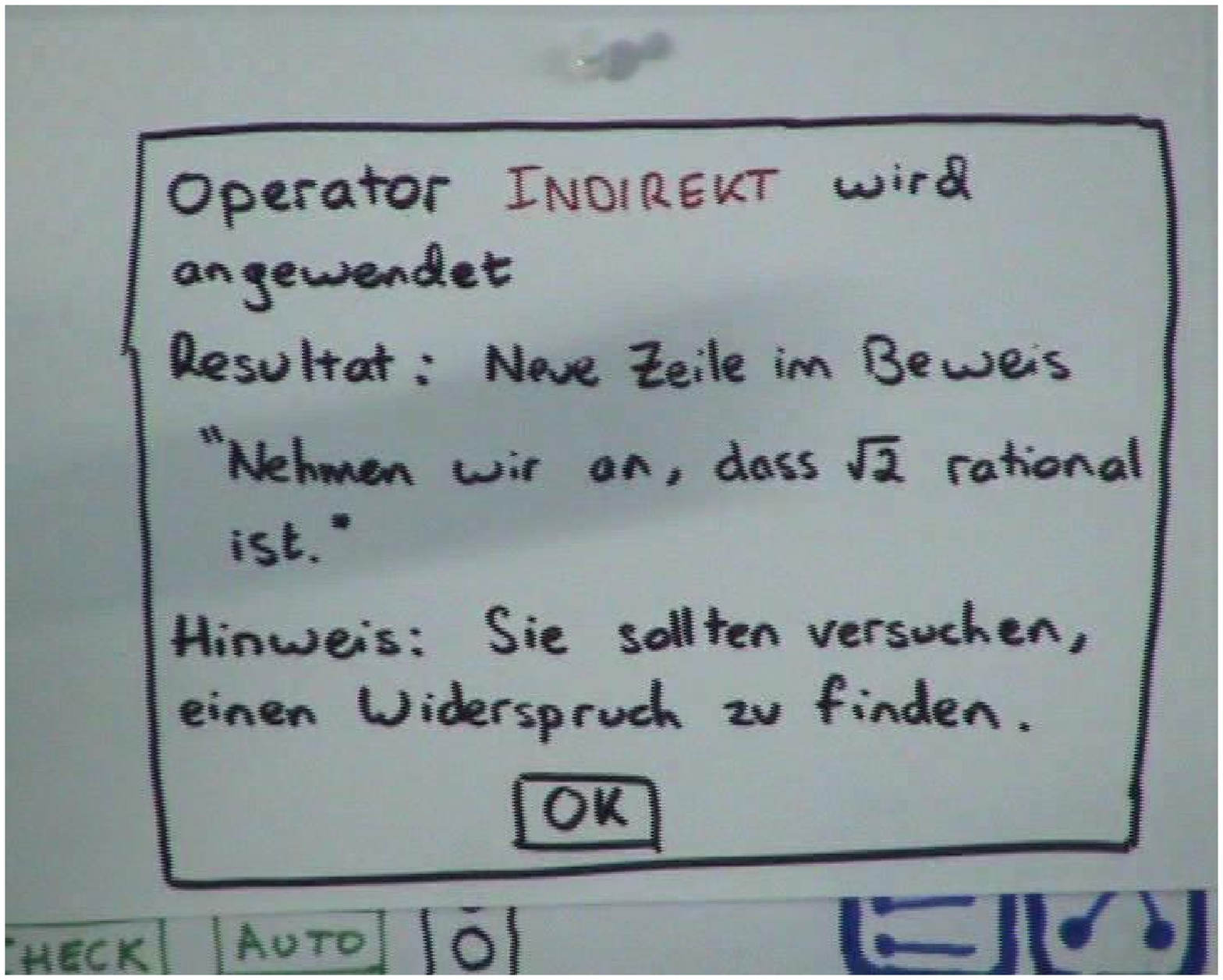}
  }
  \caption{\label{fig:team_a_operator_feedback} Operator application feedback
  (Group A)}
  \end{figure}

  \begin{figure} 
  \centering{ 
   \includegraphics[width=.8\textwidth]{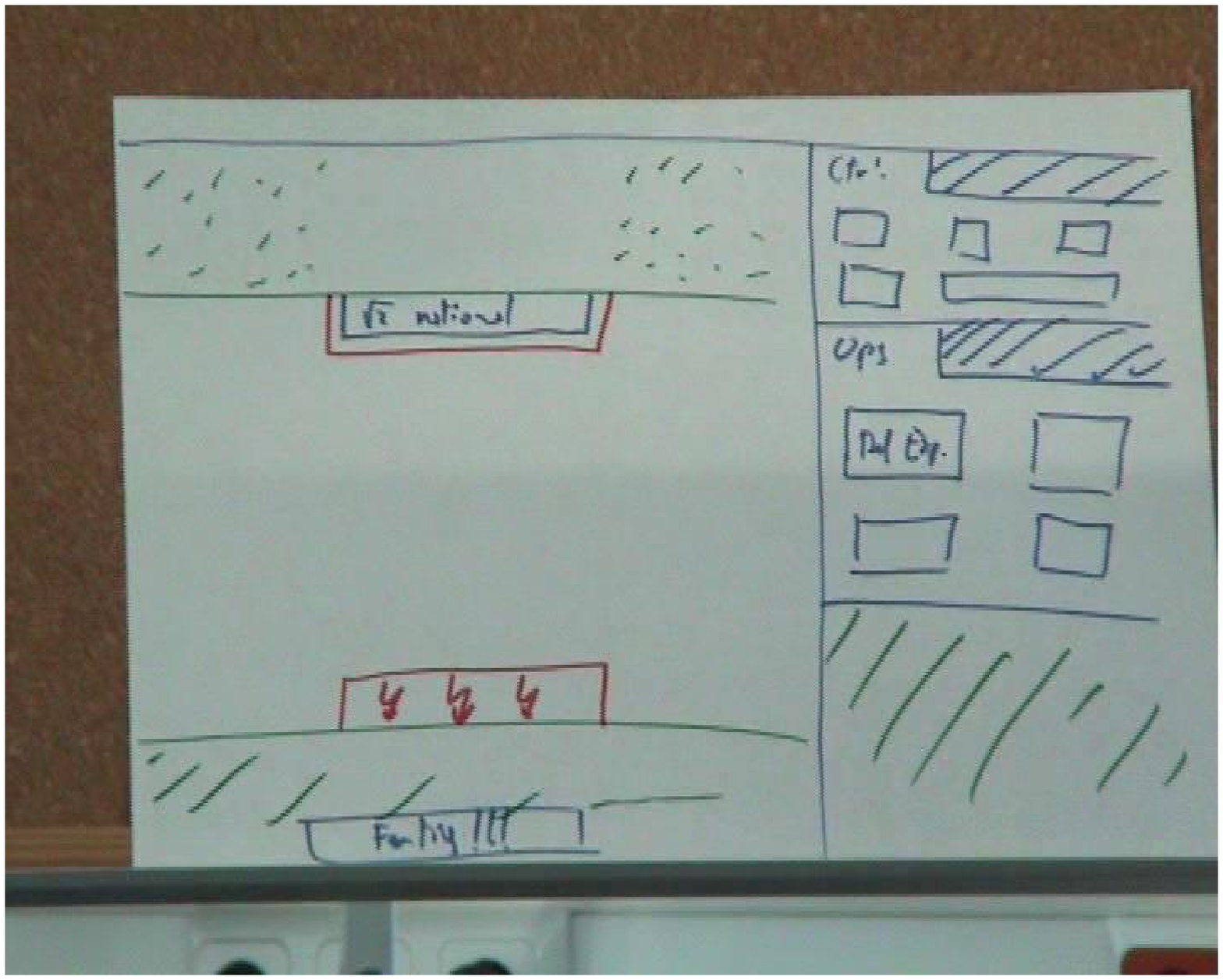}
  }
  \caption{\label{fig:team_b_window} GUI window (Group B)}
  \end{figure}

  \begin{figure} 
  \centering{ 
   \includegraphics[width=.8\textwidth]{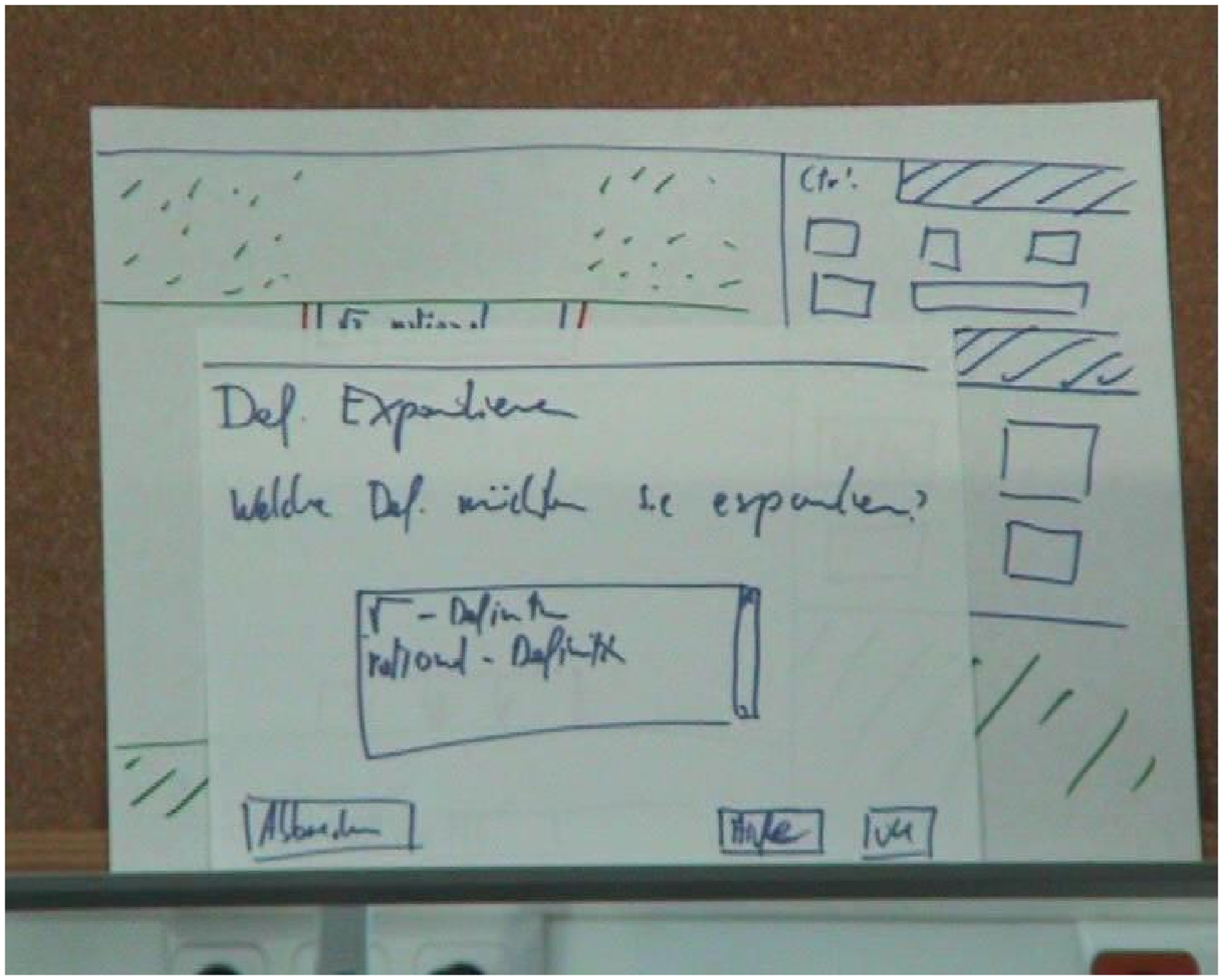}
  }
  \caption{\label{fig:team_b_operator} Operator application (Group B)}
  \end{figure}

  \begin{figure} 
  \centering{ 
   \includegraphics[width=.8\textwidth]{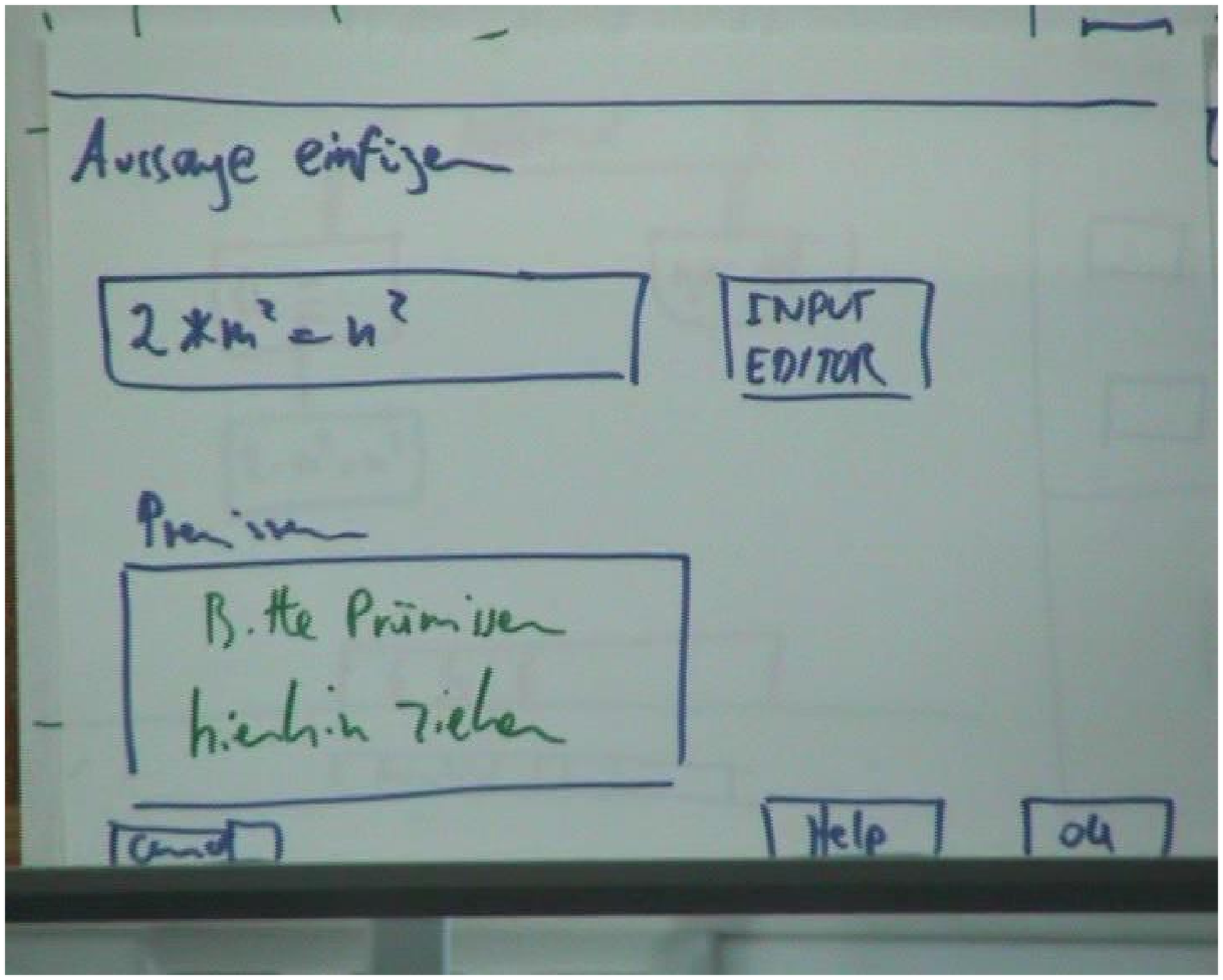}
  }
  \caption{\label{fig:team_b_island} Island insertion (Group B)}
  \end{figure}

  \begin{figure} 
  \centering{ 
   \includegraphics[width=.8\textwidth]{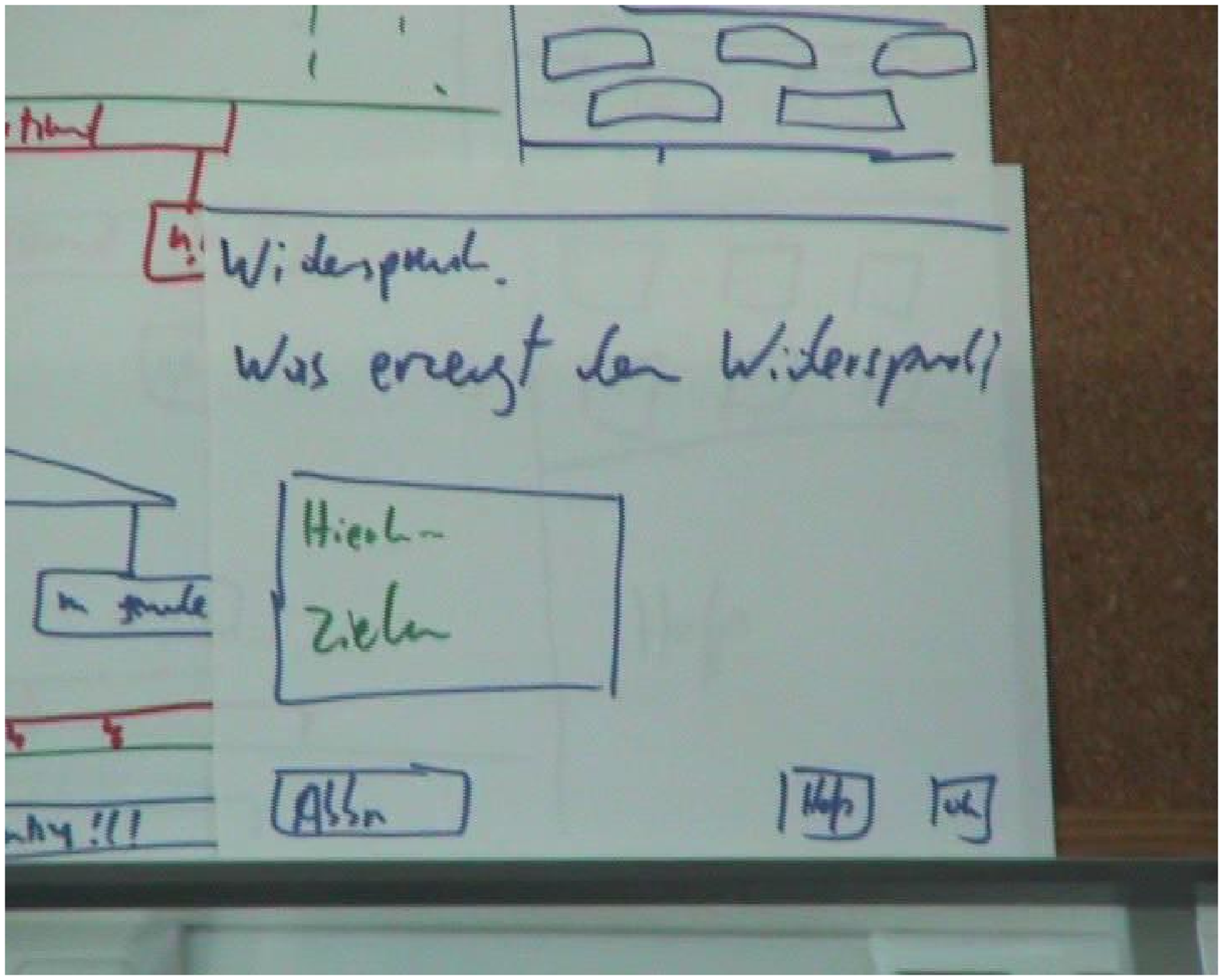}
  }
  \caption{\label{fig:team_b_contradiction} Contradiction (Group B)}
  \end{figure}

  \begin{figure} 
  \centering{ 
   \includegraphics[width=.8\textwidth]{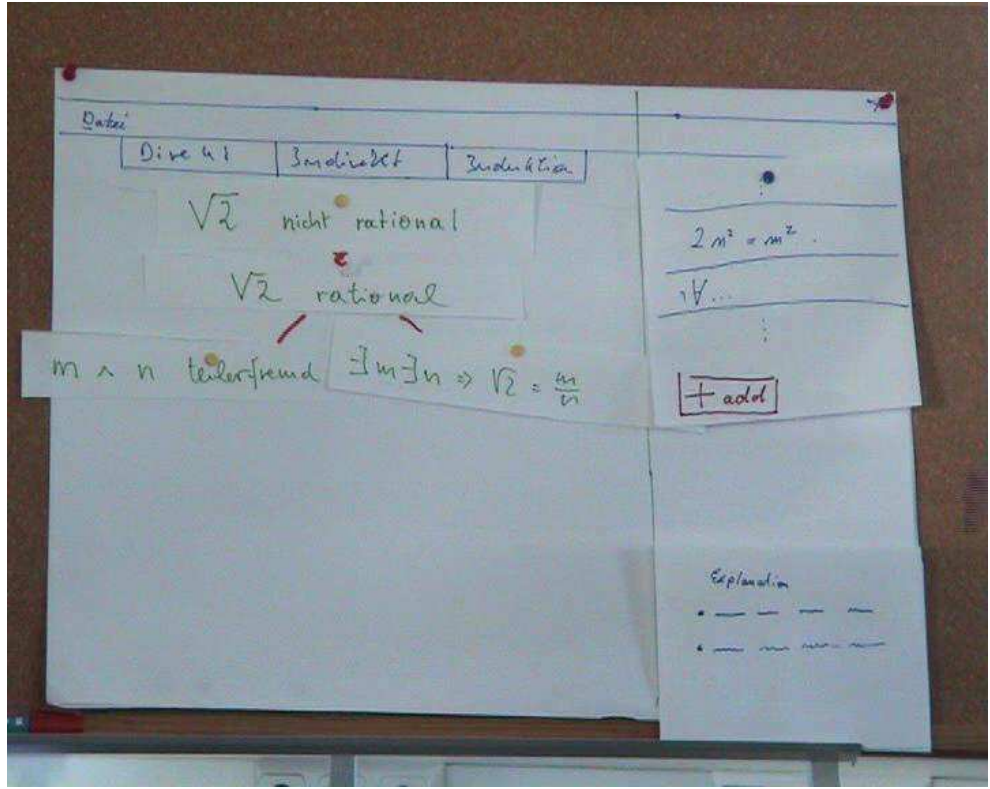}
  }
  \caption{\label{fig:team_c_window} GUI window (Group C)}
  \end{figure}

  \begin{figure} 
  \centering{ 
   \includegraphics[width=.8\textwidth]{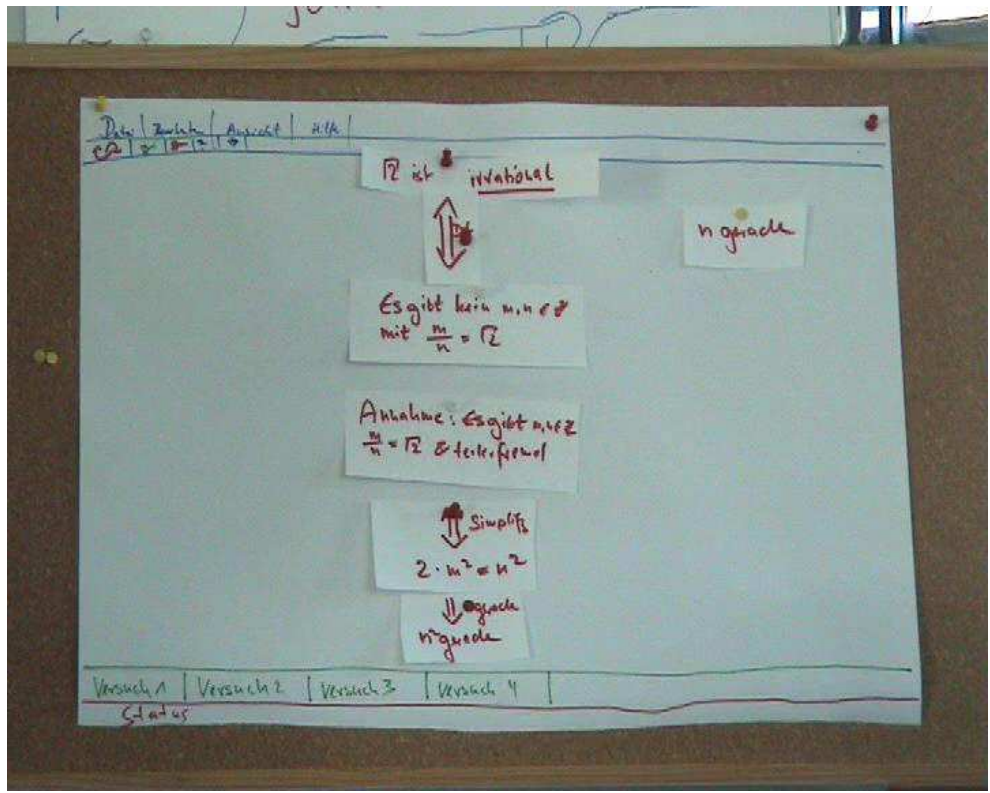}
  }
  \caption{\label{fig:team_d_window} GUI window (Group D)}
  \end{figure}

  \begin{figure}
  \centering{ 
   \includegraphics
      [width=.8\textwidth]{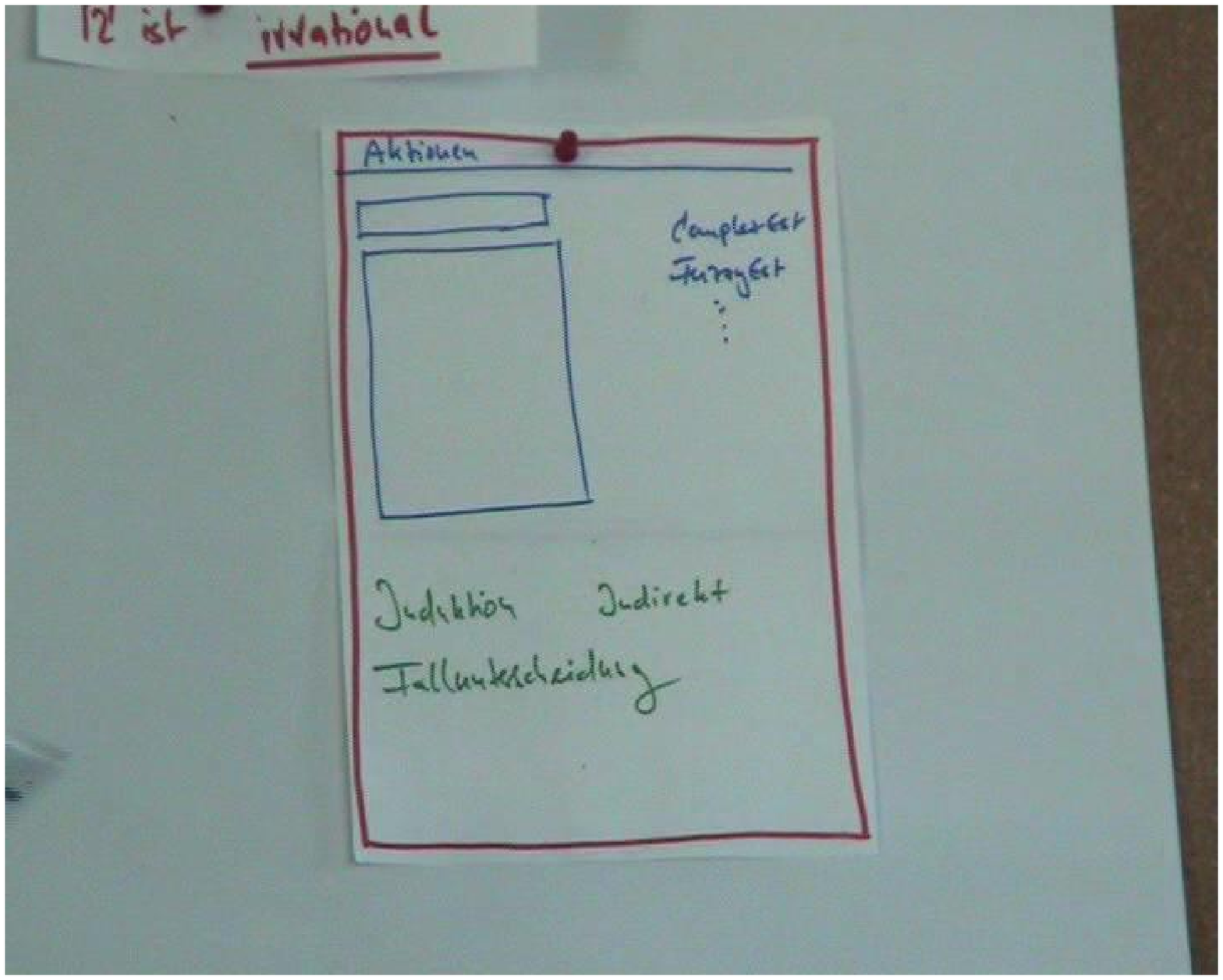}}
  \caption{\label{fig:team_d_operator} Operator (Group D)}
  \end{figure}
\end{appendix}
\end{document}